\documentclass[11pt]{article}  
\usepackage[preprint]{acl}  
\usepackage{microtype}  
\usepackage{graphicx}  
\usepackage{amsmath}  
 \usepackage{amssymb}
 
\usepackage{fvextra} 
\DefineVerbatimEnvironment{wrappedverb}{Verbatim}  
{breaklines=true, breakanywhere=true, fontsize=\tiny}  
  
\usepackage[dvipsnames]{xcolor}  

\usepackage{xcolor}
\usepackage{xparse}
\usepackage[normalem]{ulem}

\newif\ifcomments
\commentstrue        

\NewDocumentCommand{\cmt}{O{red} m}{%
  \ifcomments
    {\color{#1}%
      [\ignorespaces #2\unskip]%
    }%
  \fi
}

\usepackage{enumitem}  

\title{Fast and Effective On-policy Distillation from Reasoning Prefixes}  
\author{  
  Dongxu Zhang$^{\ast}$ \quad Zhichao Yang \quad Sepehr Janghorbani \quad Jun Han \\
  \textbf{Andrew Ressler II \quad Qian Qian \quad Gregory D. Lyng } \\  
  \textbf{ Sanjit Singh Batra$^{\ast}$ \quad Robert E. Tillman\thanks{Correspondence to:\newline 
  \texttt{\{{dongxu.zhang,sanjit.batra,rob.tillman}\}@optum.com}
}}\\
Optum AI
}

\begin{document}  
\maketitle  
  
\begin{abstract}  

On-policy distillation (OPD), which samples trajectories from the student model and supervises them with a teacher at the token level, avoids relying solely on verifiable terminal rewards and can yield better generalization than off-policy distillation. However, OPD requires expensive on-the-fly sampling of the student policy during training, which substantially increases training cost, especially for long responses. Our initial analysis shows that, during OPD, training signals are often concentrated in the prefix of each output, and that even a short teacher-generated prefix can significantly help the student produce the correct answer. Motivated by these observations, we propose a simple yet effective modification of OPD: we apply the distillation objective only to \emph{prefixes} of student-generated outputs and terminate each sampling early during distillation. Experiments on a suite of AI-for-Math and out-of-domain benchmarks show that on-policy prefix distillation matches the performance of full OPD while reducing training FLOP by 2×–47×.
\end{abstract}

\section{Introduction}  
As large language models (LLMs) continue to scale in size, model compression~\cite{buciluǎ2006model} becomes increasingly important for deploying LLMs to real-world applications where resources are often limited.  
Knowledge distillation~\cite{hinton2015distilling} is a widely-used technique in machine learning for compressing a large teacher model into a smaller student model while retaining the teacher's capabilities.  
  
In the context of LLMs, distillation is commonly performed via  
\emph{sequence-level knowledge distillation} (SeqKD), where the teacher first generates a fixed set of off-policy token trajectories and the student is trained via next-token prediction on these teacher outputs (i.e., SFT on teacher-generated sequences)  
\cite{kim2016sequence,radford2019language,wei2022finetuned,ouyang2022training}.  
Despite its broad adoption, this approach has a notable limitation, known as the \emph{exposure bias problem}, where the training is exclusively using the teacher-generated data; therefore, the student rarely learns from its own induced states and errors can accumulate at test time~\cite{lin-etal-2020-autoregressive}.  
More broadly, recent work suggests that aggressive SFT can be associated with degraded generalization~\citep{chu2025sft}, catastrophic forgetting~\citep{luo2025empirical,kalajdzievski2024scaling}, and increased hallucination rates~\citep{gekhman2024does,kalai2025language}, highlighting the need for training signals that better match the student’s deployment-time state distribution.  
These limitations also motivate moving beyond purely off-policy imitation.  
  
On-policy learning~\cite{williams1992simple}, widely studied in reinforcement learning (RL), trains a model using trajectories sampled from its own policy, reducing train--test mismatch.  
In LLM training, this means sampling from the \emph{current} student during optimization rather than relying on a fixed teacher-generated dataset.  
However, many on-policy RL methods require full rollouts to obtain terminal rewards~\cite{tulu3}, which can be prohibitively expensive for long-form reasoning.  
  
On-policy distillation (OPD)~\citep{lin-etal-2020-autoregressive,gu2024minillm,Ko2024DistiLLMTS,agarwal2024policy,lu2025onpolicydistillation} addresses key limitations of these issues from SFT and RL by minimizing the divergence between policies of student-generated outputs (SGO) and a teacher at token level.  
Compared to reward-based RL, OPD provides dense, token-level teacher feedback without requiring task-specific reward design.  
This dense supervision can make OPD more learning-efficient than sparse-reward RL; however, each update is still expensive:  
to preserve the on-policy property, OPD must generate student rollouts on the fly at each update rather than reusing a static off-policy dataset.  
In reasoning-heavy tasks, these solution trajectories can span tens of thousands of tokens, making both sampling and teacher scoring prohibitively expensive.  
  
\begin{figure}[t]  
    \centering  
    \includegraphics[width=0.45\textwidth]{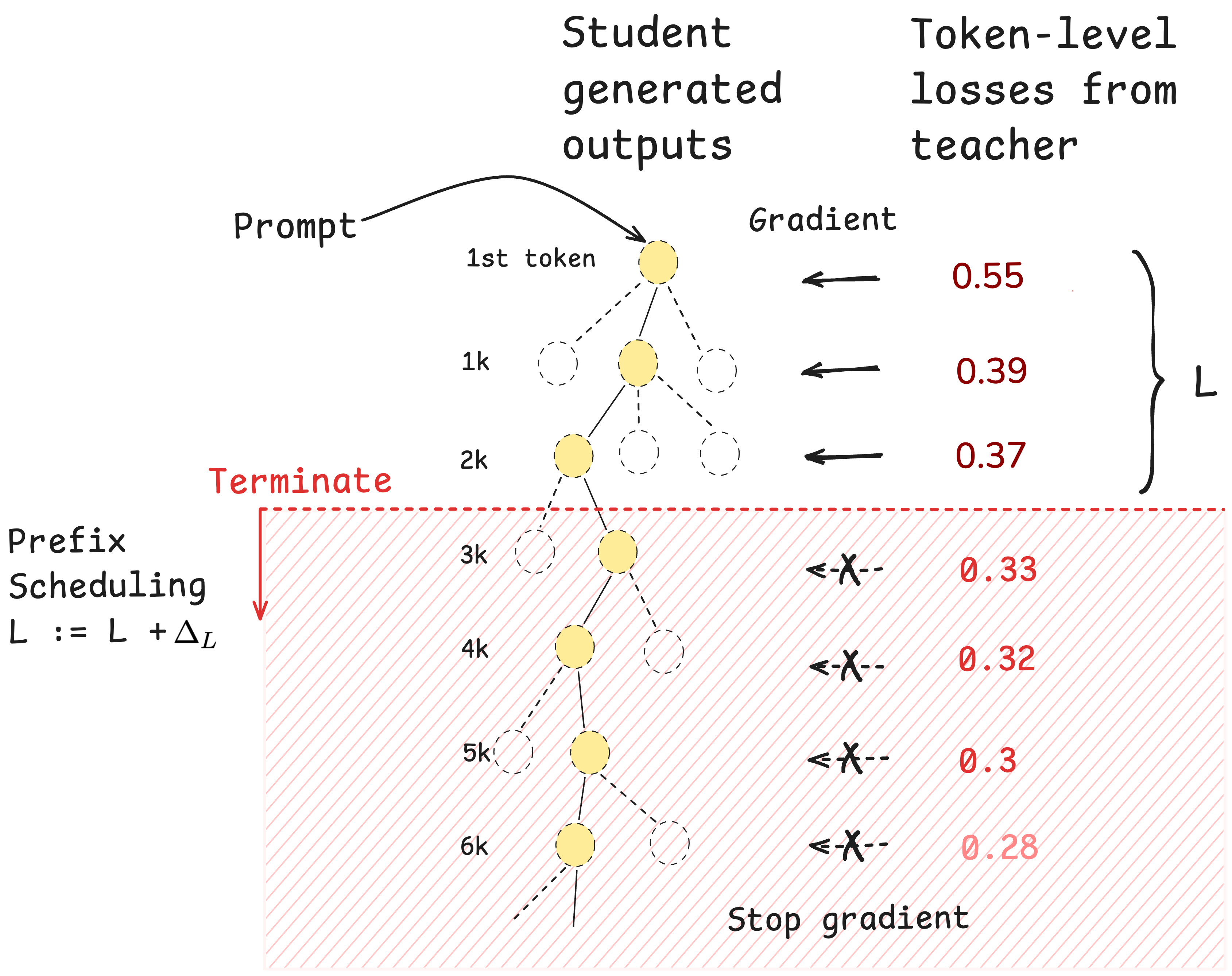}\hfill  
    \caption{\textbf{On-policy prefix distillation with prefix scheduling.} During training, gradients are backpropagated only through the first $L$ tokens of each student rollout, and sampling is terminated after $L$. Prefix scheduling increases $L$ by a fixed amount at each training step. At test time, the student generates autoregressively until producing the final answer.}  
    \label{fig:story}  
\end{figure}  
  
Because OPD provides token-level teacher feedback, full-length supervision may be unnecessary.  
This raises our central research question:  
\noindent \textbf{Can we preserve the benefits of on-policy distillation while \emph{only} supervising the prefix of each student trajectory, thereby reducing the training cost?}  
  
We test two hypotheses: (a) OPD’s learning signal is assigned unequally towards early tokens; (b) aligning this prefix is often sufficient for the student to complete the remaining reasoning correctly.  
Section~\ref{sec:preliminary} supports these hypotheses, and experiments on math-reasoning benchmarks show that prefix OPD achieves performance close to full OPD while reducing training compute significiantly.  
  
\paragraph{Contributions.}  
\begin{itemize}[leftmargin=*, topsep=2pt, itemsep=2pt]  
    \item We propose \emph{on-policy prefix distillation}, which truncates rollouts during training and applies distillation only on the prefix.  
    \item We introduce a simple \emph{prefix scheduling} strategy that progressively increases the trained prefix length over training.  
    \item On AI-for-Math and out-of-domain benchmarks, prefix OPD closely matches full OPD while reducing training FLOP by 2×–47×.  
\end{itemize}  
  
\section{Preliminary Study} 
\label{sec:preliminary}

In this section, we first describe the  learning objective of on-policy distillation. Then, we provide experimental analysis to validate the idea of on-policy prefix distillation. 

\subsection{On-policy Distillation}

On-policy distillation generalizes classical off-policy knowledge distillation where at each iteration the student policy $\pi_s$ gathers data and is updated to maximize the similarity to a fixed teacher $\pi_T$ using Kullback–Leibler divergence: 
$$
\min_{\pi_s}\;\Bigl\{   
\mathrm{KL}(\pi_s\,\|\,\pi_T)
\Bigr\}. 
$$

Since the expectation is taken over the student’s own state distribution, OPD avoids the distributional mismatch that plagues off-policy imitation and yields gradients aligned with the deployment policy.    

We choose to use the reverse KL divergence because it encourages the student to assign high probability to teacher-preferred actions, reducing opportunities for reward hacking, which often happens when using forward KL. Its mode-seeking nature also reduces the risk of spreading the student policy over many suboptimal options \cite{gu2024minillm,he2025nondeterminism}~\footnote{We use AI4Math benchmarks in this work for evaluation, and often a math problem can be solved by more than one solution path. For example, in Appendix~\ref{appendix: aime2024 stats}, we show that AIME-2024 has approximately 5.4 unique solutions per question.}. 
In addition, though previous work \cite{agarwal2024policy} also allows using a mixture of forward and reverse KL, it shows that for arithmetic reasoning, such Jensen-Shannon divergence could lead to decreased performance.

During training, since it is intractable to sample all possible responses, we adopt the following unbiased estimation via sampling:

\begin{align}  
\mathrm{KL}(\pi_s\,\|\,\pi_T)  
= & \mathbb{E}_{x_{1:L}\sim \pi_s}\!\left[  
\sum_{t=1}^{L-1}  
\log \frac{\pi_s(x_{t+1}\mid x_{1:t})}{\pi_T(x_{t+1}\mid x_{1:t})}  
\right] \notag\\  
\approx & \frac{1}{N}\sum_{i=1}^{N}\sum_{t=1}^{L_i-1}   
\log \frac{\pi_s(x^{(i)}_{t+1}\mid x^{(i)}_{1:t})}{\pi_T(x^{(i)}_{t+1}\mid x^{(i)}_{1:t})},  
\label{eq:loss}  
\end{align}  
\noindent where $L$ is the maximum response length and $N$ is the number of samples. Note that this also allows one to avoid storing logits over the full vocabulary for each token in the trajectory~\footnote{A single sample of response with 16k tokens would take 4.8GB memory under BF16. This was not a challenge in the previous on-policy distillation work such as GKD, MiniLLM, DistillLM as the max response length was often 512.}.

Optionally, a student can also receive rewards from the environment and/or use entropy loss or KL regularization to its old policy for robustness and specific downstream needs. However, in this work we do not explore these losses for simplicity and focus on the on-policy distillation component.

Note that we leverage vllm~\cite{kwon2023efficient} to speed up sampling process. To correct the off-policy issue due to floating-point non-associativity during rollouts~\cite{he2025nondeterminism}, 
we adopt importance sampling during gradient update~
\footnote{For more details, please refer to ~\url{https://tinker-docs.thinkingmachines.ai/losses\#policy-gradient-importance_sampling}. We also follow this implementation and sum the token-level losses over the sequence length.}.

\begin{figure}[t]  
    \centering  
    \includegraphics[width=0.5\textwidth]{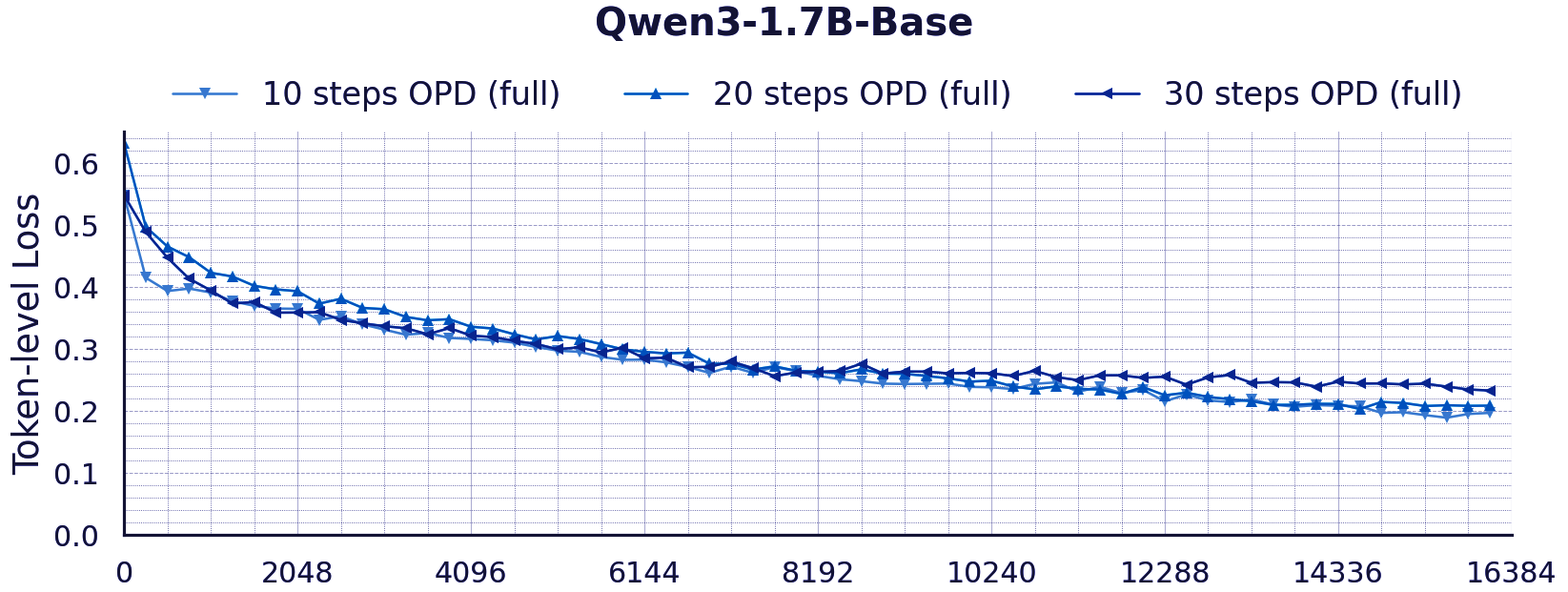}  
    \includegraphics[width=0.5\textwidth]{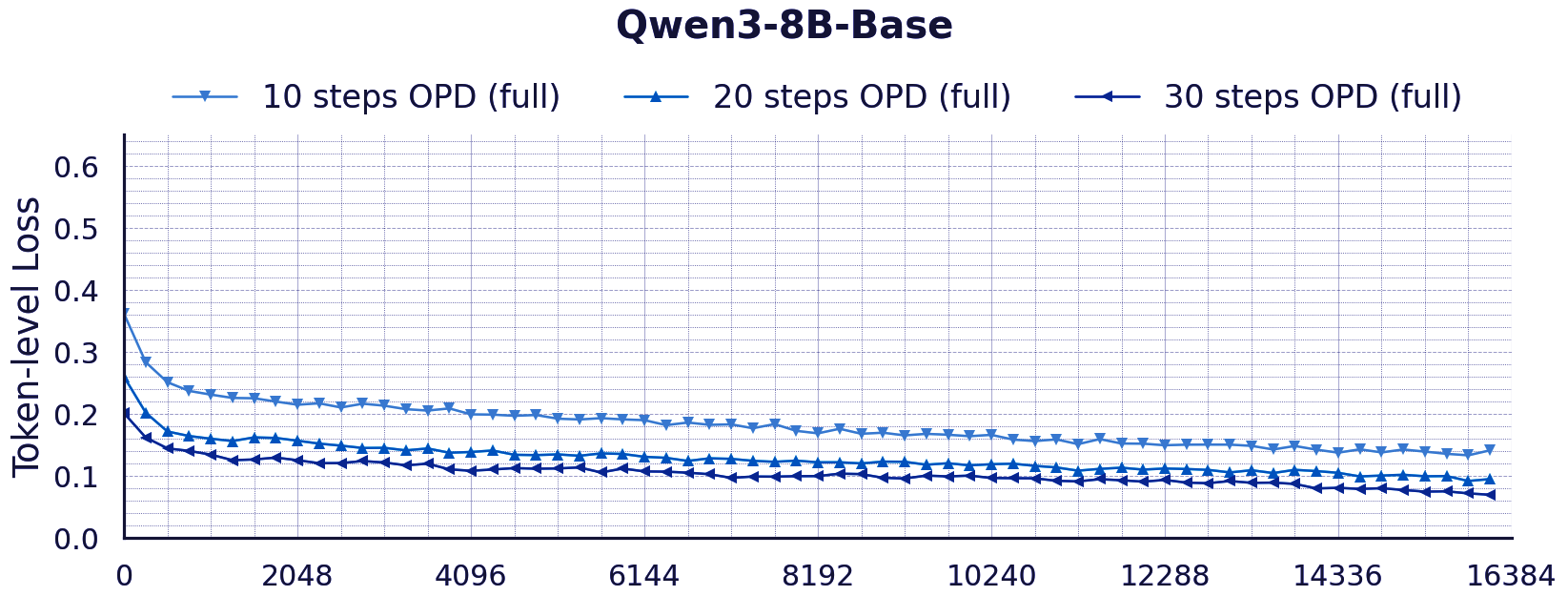}  
    \caption{Distribution of reverse-KL loss along the output position during full on-policy distillation over OpenThoughts3. \textbf{Top}: Student model is Qwen3-1.7B-Base. \textbf{Bottom}: Student model is Qwen3-8B-Base. }  
    \label{fig:advantage_distribution}  
\end{figure}  

\subsection{Does Prefix Receive More Training Signals? }

To understand where the learning signal originates from, we measure the token-level average reverse-KL loss from Eq.~\ref{eq:loss} over tokens grouped into each positional bin along the trajectory during on-policy distillation training.

We use Qwen3-1.7B-Base and Qwen3-8B-Base~\cite{yang2025qwen3} as the student models, and Qwen3-8B as the teacher model. 
Figure~\ref{fig:advantage_distribution} shows the distribution of the average magnitude of reverse-KL loss at 10th, 20th, and 30th steps of full on-policy distillation training on the questions sampled from OpenThoughts3~\citep{guha2025openthoughts} dataset. Each point in the curve corresponds to a positional bin covering 256 tokens, with a maximum output length of 16,384 tokens. 
During training, batch size 512 and learning rate 5e-5 are used, following ~\citet{lu2025onpolicydistillation}.

The distribution plot reveals that training signal is concentrated in the earlier generated tokens, and this effect is consistent across various training steps and model sizes. We conjecture that this pattern reflects the student model being weaker at high-level planning captured early in the trajectory than at executing later steps of a plan.

\subsection{Can a Student Model Complete a Truncated Response from the Teacher without Training?}  
  
Another question is whether a student that receives supervision only at the beginning of its responses can subsequently finish the solution correctly.    
To probe this, we conduct a \emph{teacher-prefix} study: we let the teacher LLM (Qwen3-8B) generate full solutions, feed the first $L$ tokens ($L\!\in\!\{2048,4096,6144, 8192\}$) to un-trained students, and ask them to continue the reasoning unaided.    
This experiment is purely probing the model’s ability to “fill in the tail.”

\begin{figure}[t]  
    \centering  
    \includegraphics[width=0.4\textwidth]{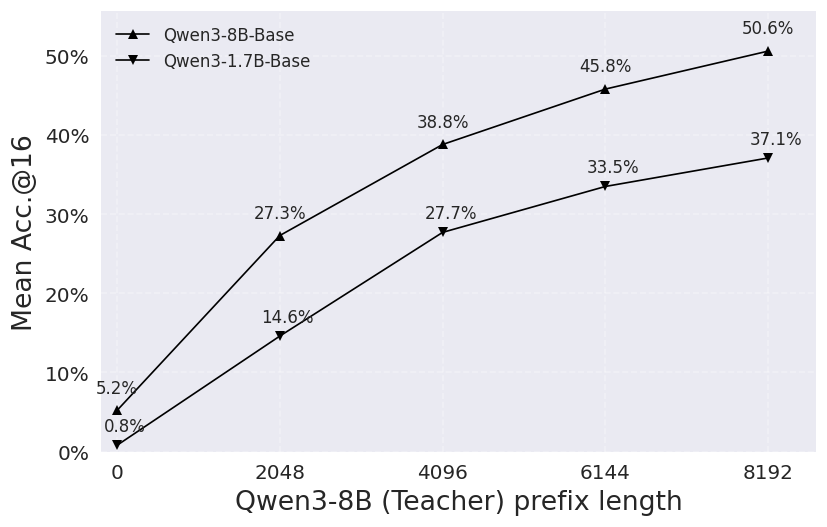}  
    \caption{\textsc{Mean Acc.@16} on AIME-24 when an untrained student (Qwen3-8B-Base and Qwen3-1.7B-Base) continues from varying lengths of a teacher (Qwen3-8B)-generated prefix. 
    }  
    \label{fig:teacher_prefix_and_base_model}  
\end{figure}  

Figure~\ref{fig:teacher_prefix_and_base_model} reports mean accuracy at 16 samples on AIME-24. During this experiment, we also ensure that teacher's final answer is not likely in the prefix by enforcing  $L_{p} \leftarrow \min\{L_{p}, L_{r} - 512\}$ where $L_{p}$ is the length of teacher's prefix and $L_{r}$ is the length of teacher's full response for the current input. Note that teacher’s average output length on AIME-24 is 15k and maximum output length for this experiment is set to 32k.

Firstly, the plot shows that teacher's reasoning prefix does improve students' performances, suggesting that the early guidance from the teacher can effectively help students to infer the remaining steps even without further training.
In addition, the curve also shows that the performance jumps sharply given the first few thousands teacher tokens and the increase slows down thereafter. This again verifies that the prefix is crucial during the reasoning process. We further provide a case study from AIME-24 to illustrate the effect of the teacher's reasoning prefix in Section~\ref{sec:case study}.

\section{Methodology}  

Figure~\ref{fig:story} shows the main idea of prefix distillation.
In this work, we enforce prefix-only training by limiting the maximum number of tokens from generated response during \emph{training} to $L^{\text{train}}$. 
No additional modification to the loss or sampling procedure is required; we simply early-stop the on-policy generation once the budget is reached, back-propagate through the truncated trajectory for gradient updates, then proceed to the next batch. The on-policy property is preserved for the first $L^{\text{train}}$ tokens.

At evaluation time we remove the cap\footnote{In the rest of the experiments, we allow up to $L^{\text{test}}{=}16384$ tokens}, ensuring that final task performance is not artificially constrained.    
This design choice keeps the implementation minimal, yet yields significant savings of computation during training. 
  
\paragraph{Special Tokens.} Though memory friendly for long reasoning scenarios, we observe that it is challenging for the \emph{sample-based} KL estimation in Eq.~\ref{eq:loss} to learn from generating special tokens. When such supervision is provided from the teacher, the student repeatedly attempts to switch to a different token which leads to generating unreadable content. To avoid falling into such a scenario, in our training, \texttt{<think>} is enforced at the end of the input prompt, to align with the teacher's behavior.

\paragraph{Prefix Scheduling.}  
Although short prefixes reduce cost, later tokens in the student's rollout can still carry valuable supervision when scored under the teacher. To progressively expose the student to longer-horizon feedback, we explore a simple linear schedule that increases the training prefix length after each gradient update:  
$  
L^{\text{train}} \leftarrow L^{\text{train}} + \Delta_L  
$.
In our experiments, $L^{\text{train}}$ starts with 1 and $\Delta_L = 256$.

\section{Experiments}

\subsection{Setup}

\begin{table*}[h!]  
\centering  
\small  
\begin{tabular}{l|r|r|r||r|r||r|r}  
\hline  
 & \multicolumn{3}{c||}{In-Domain}   & \multicolumn{2}{c||}{Out-domain} &\multicolumn{2}{c}{Training cost}  \\ 
\hline  
 & {\scriptsize MATH500} & {\scriptsize AIME-24} & {\scriptsize AIME-25}   & {\scriptsize GPQA} & {\scriptsize MMLU-Pro} & FLOP & Hours \\  

\hline  
Qwen3-1.7B-Base    & 20.1 &  0.2 &  0.0 & 11.2  & 9.1 & \multicolumn{2}{c}{No post training}  \\  
\hline
\multicolumn{8}{c}{\emph{w. limited budgets}} \\
\hline
OPD - 10 steps      & 43.4 & 5.4 & 3.5 & \underline{18.3} & \textbf{30.7} & $8.2\times 10^{18}$ &  12.2 \\ 
SeqKD - 300 steps          & \textbf{60.5} & 4.6 & \underline{6.9} & \textbf{20.2} &  \underline{28.5} & $1.3\times 10^{19}$  & 3.7 \\ 
\hline
\hline
OPD (prefix 256)       & 50.0 & 4.8 & 6.3  & 1.0 & 1.5 & $9.4\times 10^{17}$ & 2.4 \\ 
OPD (prefix 512)       & 47.6 & 4.2  & 6.0 & 1.1 & 1.0 & $1.4\times 10^{18}$  & 2.8 \\ 
OPD (prefix 1024)      & 54.5 & \textbf{8.1} & \textbf{9.0} & 1.1 & 1.9 & $2.5\times 10^{18}$ & 3.4 \\ 
OPD (prefix 2048)      & \underline{58.3} & \underline{7.3} & 5.8 & 4.5 & 12.0 &  $4.7\times 10^{18}$ & 5.0 \\ 
\hline
\hline
\multicolumn{8}{c}{\emph{w. sufficient budgets}} \\
\hline
OPD   & 67.3 & 10.0  & 11.5 & 24.6 & 41.3 & $5.7\times 10^{19}$ & 68.6  \\ 
SeqKD        & 70.8 & 8.8 & 14.4  &  19.8 & 29.6 & $6.6\times 10^{19}$  & 18.3 \\ 
\hline
OPD (prefix scheduling)    & 68.1 & 10.6  & 11.5 & 23.6 & 39.4 & $2.4\times 10^{19}$  & 30.8 \\ 
\hline  
\end{tabular}  
\caption{Mean accuracies and costs with Qwen3-1.7B-Base as the student.}
\label{tab:main result 1.7b}  
\end{table*}

\begin{table*}[h!]  
\centering  
\small  
\begin{tabular}{l|r|r|r||r|r||r|r}  
\hline  
 & \multicolumn{3}{c||}{In-Domain}   & \multicolumn{2}{c||}{Out-domain} &\multicolumn{2}{c}{Training cost}  \\ 
\hline  
 & {\scriptsize MATH500} & {\scriptsize AIME-24} & {\scriptsize AIME-25}   & {\scriptsize GPQA} & {\scriptsize MMLU-Pro} & FLOP & Hours \\  

\hline  
Qwen3-8B-Base         & 39.0 &  5.2  & 2.1 & 17.6 & 26.5 & \multicolumn{2}{c}{No post training}  \\  
\hline
\multicolumn{8}{c}{\emph{w. limited budgets}} \\
\hline
OPD - 10 steps       & 77.6 & 23.3 & 24.2  & \textbf{47.5} & \textbf{62.7} & $2.7\times 10^{19}$ & 23.3 \\  
SeqKD - 300 steps                      & \textbf{88.3} & 31.9 & 25.4 &  38.5 & 52.6 & $4.4\times 10^{19}$ & 11.3 \\  
\hline
\hline
OPD (prefix 256)      & 82.0  &  30.8 & 26.3  & 35.5 & 58.9 & $3.4\times 10^{18}$ & 3.2 \\  
OPD (prefix 512)      & 84.7 & 33.5  & 26.7 &  35.7 &  60.3 &$5.2\times 10^{18}$ & 4.8 \\  
OPD (prefix 1024)     &  84.1 &  \underline{36.5} & \underline{28.3}  & 37.1 & 60.5 & $8.9\times 10^{18}$  & 5.6 \\ 
OPD (prefix 2048)     & \underline{87.1}  &  \textbf{39.2} & \textbf{31.3}  & \underline{40.0} & \underline{61.3} & $1.7\times 10^{19}$  & 9.8 \\  

\hline
\hline
\multicolumn{8}{c}{\emph{w. sufficient budgets}} \\
\hline
OPD               & 87.4 &  44.2 & 33.5  & 48.8 & 66.5 & $1.8\times 10^{20}$ & 148.3 \\  
SeqKD                    & 91.3 &  41.7 & 34.6 & 42.2 & 55.4 & $2.2\times 10^{20}$ & 56.7 \\  
\hline
OPD (prefix scheduling)    & 86.9 & 44.0  &  33.5 & 48.3 & 66.1 & $7.5\times 10^{19}$  & 53.0 \\ 
\hline  
\end{tabular}  
\caption{Mean accuracies and costs with Qwen3-8B-Base as the student.}
\label{tab:main result}  
\end{table*}  

We randomly sample questions from \textsc{OpenThoughts3}~\citep{guha2025openthoughts} for training, use Qwen3-1.7B-Base / Qwen3-8B-Base as the student, and Qwen3-8B as the on-policy teacher. We leverage off-policy responses of QwQ-32B~\cite{qwen2024qwq} from OpenThoughts3 for SeqKD~\footnote{Note that QwQ-32B performs slightly better than Qwen3-8B on MATH-500, AIME-24 and AIME-25 based on \citet{yang2025qwen3}. }.

\noindent \textbf{Hyperparameters.} Following hyper-parameters from ~\citet{lu2025onpolicydistillation}, for SeqKD, we use batch size 128, learning rate $1\times10^{-4}$; for OPD and our prefix variants, we use batch size 512, 4 samples per prompt, learning rate $5\times10^{-5}$. For more details on the effect of different learning rates, please refer to Appendix~\ref{appendix:small learning rate}. We implement training with the \textsc{Verl} framework~\cite{sheng2024hybridflow} and optimize using AdamW.    
All OPD experiments are run for 60 training steps, capped by a 7-day budget of 8xA100 GPUs for the full OPD training using Qwen3-8B-Base. 
We use in total 30.7k questions for OPD training, and 192k data for SeqKD to reach a comparable FLOP.\footnote{OPD uses significantly higher FLOP per prompt than SeqKD because it requires multiple response sampling during training and needs to calculate $\log \pi_T(x_{t+1}\mid x_1,\dots,x_t)$. }

\noindent \textbf{Benchmarks.} 
Evaluation is performed on \textsc{MATH}500~\citep{hendrycks2measuring}, AIME-24, AIME-25 as in-domain testing,  GPQA~\citep{rein2024gpqa}, and a random sample of 500 questions from MMLU-Pro~\cite{wang2024mmlu} as out-of-domain testing. During evaluation, temperature = 1.0, top\_p = 1.0 and maximum response length is 16\,384. 

\noindent \textbf{Metrics.} 
For MATH500, GPQA and MMLU-Pro, we report mean accuracy among 4 samples. For AIME24 and AIME25 we calculate mean accuracy over 16 samples to reduce the variance. We use AIME24 as a development set and choose the best checkpoint for each baseline to report accuracy. More details can be found in Appendix~\ref{appendix: early stop}.
To compare computational cost during training, we estimate the number of GPU \textsc{FLOP} (floating point operations) including forward sampling, log prob computation from teacher and student, backward passes, and gradient updates accordingly. For more details, please refer to Appendix~\ref{appendix: flop}.
We also report GPU hours\footnote{We report hours on an 8xA100 VM. And evaluation time is excluded.} to take into account I/O, CPU and other computation cost during training. 

\noindent \textbf{Baselines.} We compare the base model, off-policy SeqKD, OPD with full rollout of maximum 16,384 tokens during training, and prefix OPD. 
We select two checkpoints as our baselines, student after 10th gradient descent of full OPD, and student with 300 steps of SeqKD. These two baselines are chosen because they have a similar order of magnitude in training cost, albeit slightly higher than prefix OPD. 
Then, we also put the results with 60 steps of full OPD and 1500 steps of SeqKD for a  sufficient budget scenario and compare them with our prefix scheduling variant.

\subsection{Main Results}

Table~\ref{tab:main result 1.7b} and \ref{tab:main result} present our main results. 
All experiments are zero-shot. 
The best and second-best results within each column are highlighted with bold and underline accordingly, excluding results with sufficient budgets.

Observing the baselines, it is apparent that given a sufficient training budget, full OPD shows better out-of-domain performance, and similar or slightly worse in-domain performance than SeqKD, indicating that on-policy distillation indeed generalizes better than off-policy distillation.

\noindent \textbf{In-domain.} 
From in-domain results, OPD with prefix 1024 / 2048 outperforms baselines with limited budgets in AIME-24 and AIME-25 with half the FLOP. On MATH500, OPD with prefix outperforms OPD (full, 10 steps), while performing only slightly worse than SeqKD (300 steps).
When using Qwen3-8B-Base as the student, prefix OPD approaches the performances with sufficient budgets while use less than 10\% the FLOP.
These results indicate that prefix OPD can effectively reduce the training cost while maintaining a competitive in-domain performance. 

\noindent \textbf{Out-of-domain.}  
We observe different out-of-domain behavior depending on student size.  
For Qwen3-8B-Base, prefix OPD improves over the base model, and performance  
tends to move closer to full OPD as we increase $L^{\text{train}}$  
(Table~\ref{tab:main result}).  
In contrast, for Qwen3-1.7B-Base, training on very short prefixes can largely reduce  
out-of-domain accuracy below the pre-trained student (Table~\ref{tab:main result 1.7b}).  
One plausible explanation is that, for small students, optimizing only early tokens  
can improve how the solution starts but worsen the model’s ability to continue reliably  
over long responses (also see \textbf{Finding 3}).  
In these cases, prefix scheduling (gradually increasing $L^{\text{train}}$ during training)  
is a more robust choice. 
As shown from Table~\ref{tab:main result 1.7b}, OPD with prefix scheduling leads to competitive out-of-domain performance.

\noindent \textbf{Scheduling.} 
Our linear prefix scheduling strategy provides benefits of both prefix distillation and full trajectory distillation. Across all settings, scheduled prefix OPD matches the upper-bound performance of full OPD while using less than 50\% of the FLOP. These results suggest that allocating compute progressively to longer prefixes is effective. It is worth exploring more adaptive schedulers (e.g., curriculum based on validation performance or KL/entropy signals) in future work.

\subsection{Ablation Study}  
  
Prefix on-policy distillation can be interpreted as an extreme form of reward shaping in which token-level advantages outside the first $L_{\text{train}}$ positions of rollouts are multiplied by zero.    
The premise is that not every token contributes equally to downstream accuracy.    
We conduct an ablation study to compare the usefulness of different positions in the rollout during on-policy distillation. 
In particular, we apply a sliding window of same length to the rollouts and only pass through policy gradients within the window. The windows we applied are [1024, 2047], [2048, 3071], [3072, 4095] and 1024 tokens on the tail.
Results in Table~\ref{tab:ablation study} show that the prefix window leads to the best results among all ablation windows, which suggests that ``earlier is more important'' is an effective heuristic. 

\begin{table}[h!]  
\centering  
\small  
\begin{tabular}{l|r|r|r}  
\hline  
Method & Step 10 & Step 20 & Step 30 \\  
\hline  
OPD (full)       & 23.3 & 40.8 & 40.0\\ 
\hline 
OPD on prefix 1024       & \textbf{26.3} & \textbf{32.5} & \textbf{34.8} \\  
OPD on $[1024,2048)$      & 14.0 & 16.0 &  16.0 \\  
OPD on $[2048, 3072)$     & 18.8 & 16.9 & 15.8 \\  
OPD on $[3072, 4096)$     & 9.4 & 19.2 &  22.5 \\  
OPD on tail 1024        & 17.9 & 23.8 &  23.8 \\ 
\hline  
\end{tabular}  
\caption{Ablation of masking strategies with Qwen3-8B-Base as student and Qwen3-8B as the teacher. We report Mean Acc.@16 on AIME-24.
}
\label{tab:ablation study}  
\end{table}

\subsection{Case Study}  
\label{sec:case study}  
  
We use an AIME-2024 example to illustrate how \emph{early} guidance can prevent compounding reasoning errors and why prefix-only supervision can be effective. In this example, we use Qwen3-8B-Base as the initial student and Qwen3-8B as the teacher model.
  
\paragraph{An AIME-2024 Problem.}  

Let $x,y$ and $z$ be positive real numbers that satisfy the following system of equations: 
$  
\log_2\!\left(\frac{x}{yz}\right)=\frac12,\qquad  
\log_2\!\left(\frac{y}{xz}\right)=\frac13,\qquad  
\log_2\!\left(\frac{z}{xy}\right)=\frac14.  
$  
Then the value of $\left|\log_2(x^4y^3z^2)\right|$ is $\tfrac{m}{n}$ where $m$ and $n$ are relatively prime positive integers. Find $m+n$.

\paragraph{A Valid Solution Sketch.}  
Let $a=\log_2 x$, $b=\log_2 y$, $c=\log_2 z$. Using  
$\log_2\!\left(\frac{x}{yz}\right)=\log_2x-\log_2y-\log_2z$,  
the system becomes  
$  
a-b-c=\tfrac12,\qquad b-a-c=\tfrac13,\qquad c-a-b=\tfrac14.  
$  
Solving yields $a=-\tfrac{7}{24}$, $b=-\tfrac{3}{8}$, $c=-\tfrac{5}{12}$, hence  
$  
\bigl|\log_2(x^4y^3z^2)\bigr|=\bigl|4a+3b+2c\bigr|=\tfrac{25}{8},  
$  
so $m+n=33$.  
  
\paragraph{Behavior Comparison:}  
\begin{itemize}[leftmargin=0pt, labelsep=0.6em, itemindent=1.2em, nosep, label=--]
  \item \textbf{Base student (incorrect).}  
  When prompted directly, the model makes an early conversion error, e.g., rewriting the system as  
  $  
  x=2^{1/4}yz,\;  
  y=2^{1/5}xz,\;  
  z=2^{1/6}xy  
  $  
  (already inconsistent with $\tfrac12,\tfrac13,\tfrac14$), then carries this mistake through and produces a confident but wrong final answer (e.g., $m+n=5$), without detecting the inconsistency.  
  
  \item \textbf{Base student + teacher prefix (correct).}  
  If we prepend a \emph{short} teacher-generated prefix (128 tokens from Qwen3-8B; containing only the initial reasoning, excluding the final answer), the same student follows the correct transformations  
  (e.g., $\frac{x}{yz}=2^{1/2}$), then leverages algebraic elimination and reaches the correct result $m+n=33$.  
  
  \item \textbf{Student after prefix OPD (correct).}  
  After on-policy prefix distillation, the student no longer requires an explicit teacher prefix: given the original question alone, it immediately introduces $(a,b,c)$, solves the linear equations, computes  
  $4a+3b+2c=-\tfrac{25}{8}$, and returns $m+n=33$.  
\end{itemize}  
  
This case study highlights that aligning the \emph{start} of the student’s reasoning can prevent early, compounding errors and change the final outcome. Full prompts and generations are provided in Appendix~\ref{appendix: example}.

\section{Discussion}  

\subsection{Cost and Performance Trade-off }

\begin{figure}[t]  
    \centering  
        \includegraphics[width=0.49\textwidth]{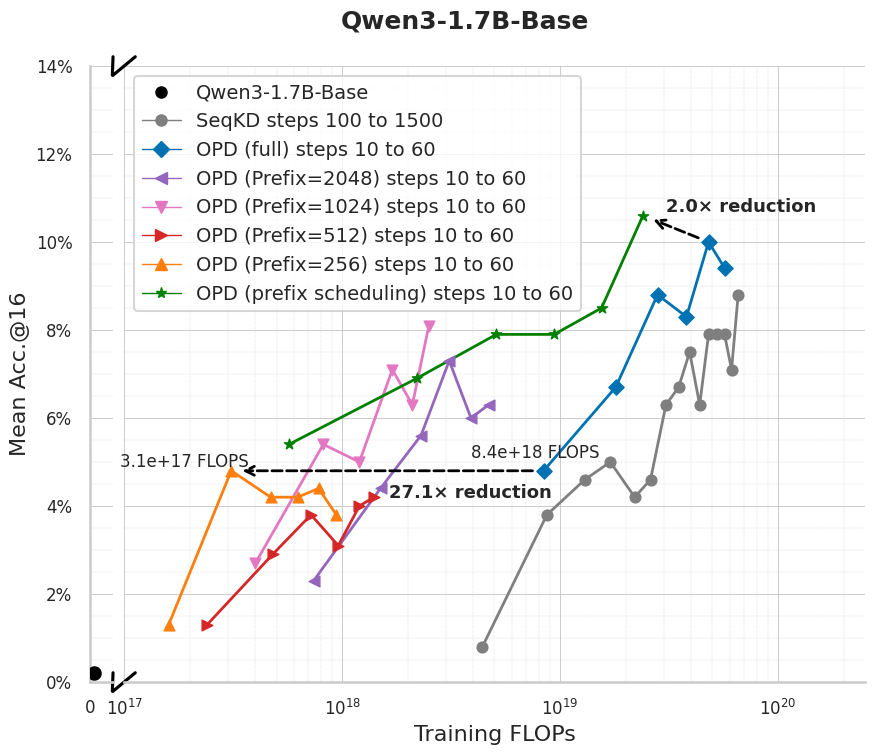}  
    \includegraphics[width=0.49\textwidth]{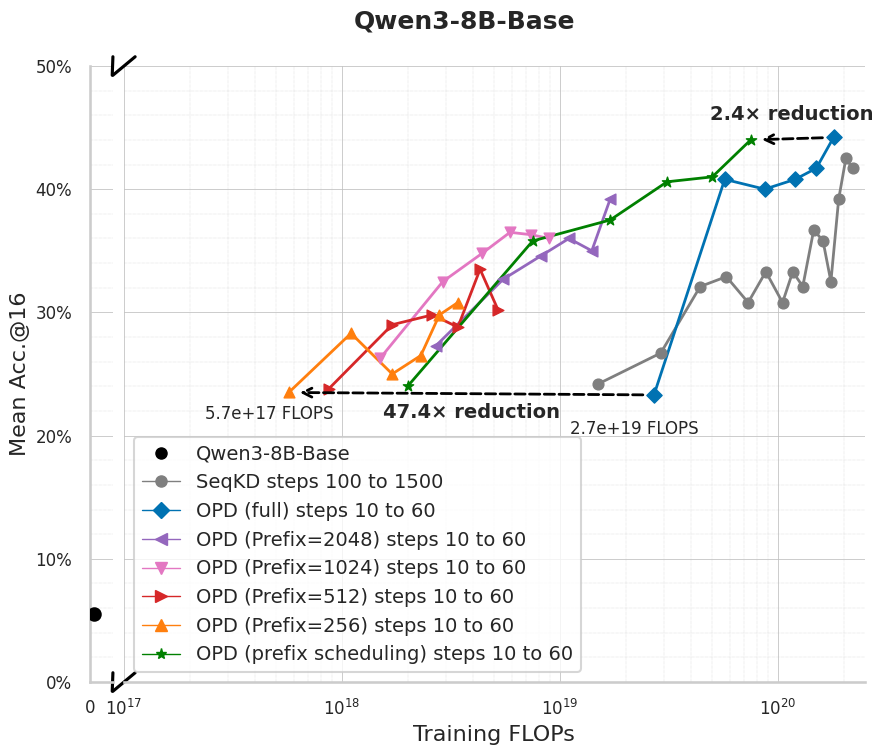} 
    \caption{Comparison of training compute vs Mean Acc.@16 on AIME-24. \textbf{Top}: Student is Qwen3-1.7B-Base. \textbf{Bottom}: Student is Qwen3-8B-Base. 
    }  
    \label{fig:main_result}  
\end{figure}

In this section, we would like to discuss the trade-off between the training cost and validation accuracy given different prefix lengths. 
Figure~\ref{fig:main_result} shows the trade-off dynamics during training on AIME-24. 

\noindent \textbf{Finding 1: Given a limited computing budget, distillation with a shorter prefix can enable significant performance gain. When the budget is sufficient, a longer prefix eventually performs better via prefix scheduling. }
If we compare performances on the left sides (10th and 20th steps) of each curve, we can see that shorter prefix can often enable significant accuracy gain in comparison to a longer prefix.  
For instance, the bottom figure shows that SeqKD requires $1.5\times 10^{19}$ to reach a 22.5\% accuracy~\footnote{For SeqKD, we didn't take into account the cost of generating the off-policy training data. In the real scenario, if we take this overhead into account, the cost could be significantly higher.}, and full OPD requires $2.7\times 10^{19}$ FLOP to reach a similar performance. In comparison, 10 steps of OPD on 256 tokens is able to increase the student model's accuracy from 5\% to 23.5\% with only $5.7 \times 10^{17}$ FLOP, a 47x reduction.

In contrast, if we compare accuracies of right sides (50th and 60th steps) of each curve, we often see that a longer prefix can lead to higher accuracies~\footnote{Exceptions are often due to variances during training.}. This suggests that when given sufficient budgets for training, a longer prefix is preferred for better performances, and that \emph{prefix scheduling} is highly recommended rather than a full distillation from the beginning of training. This strategy can often save at least half of the training cost.

\subsection{How Does Prefix Learning Affect the Tail?}

To investigate the effect of prefix-only learning beyond the prefix length, we plot a token level loss distribution with OPD with prefix=1024 on Qwen3-8B-Base on Openthoughts3. 

\begin{figure}[t]  
    \centering  
    \includegraphics[width=0.5\textwidth]{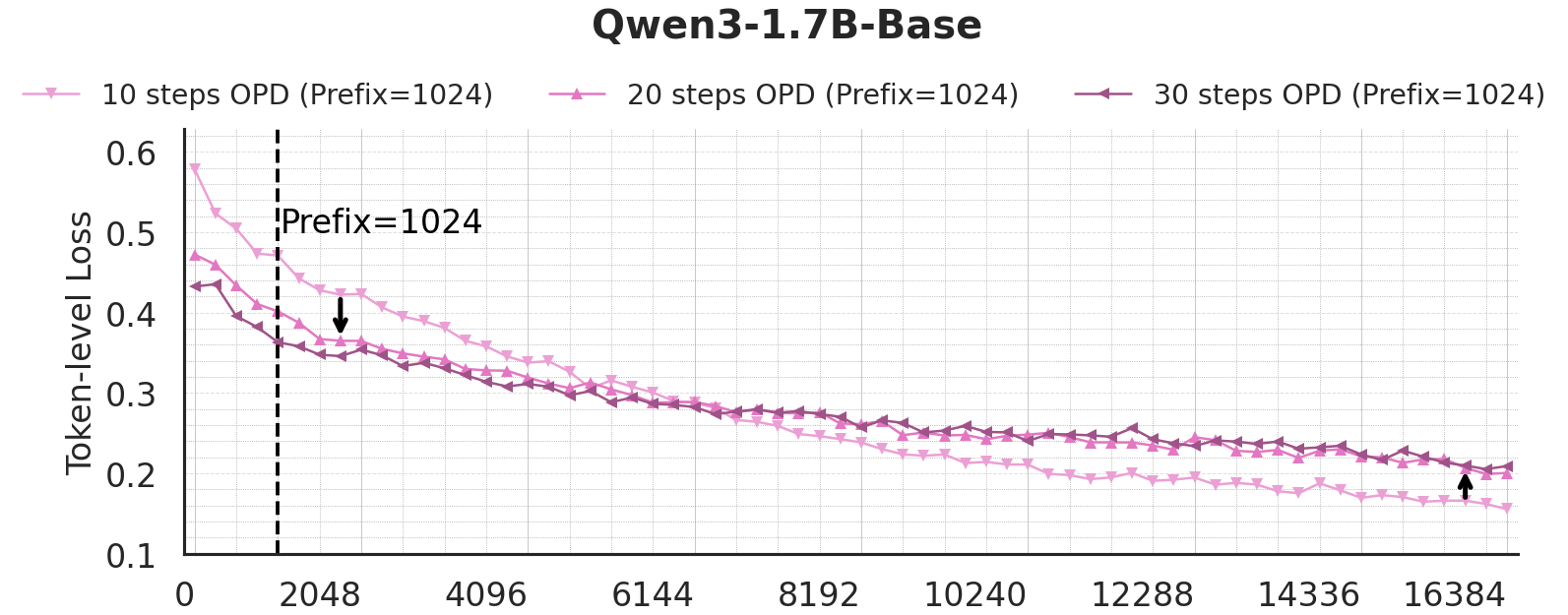}  
   \includegraphics[width=0.5\textwidth]{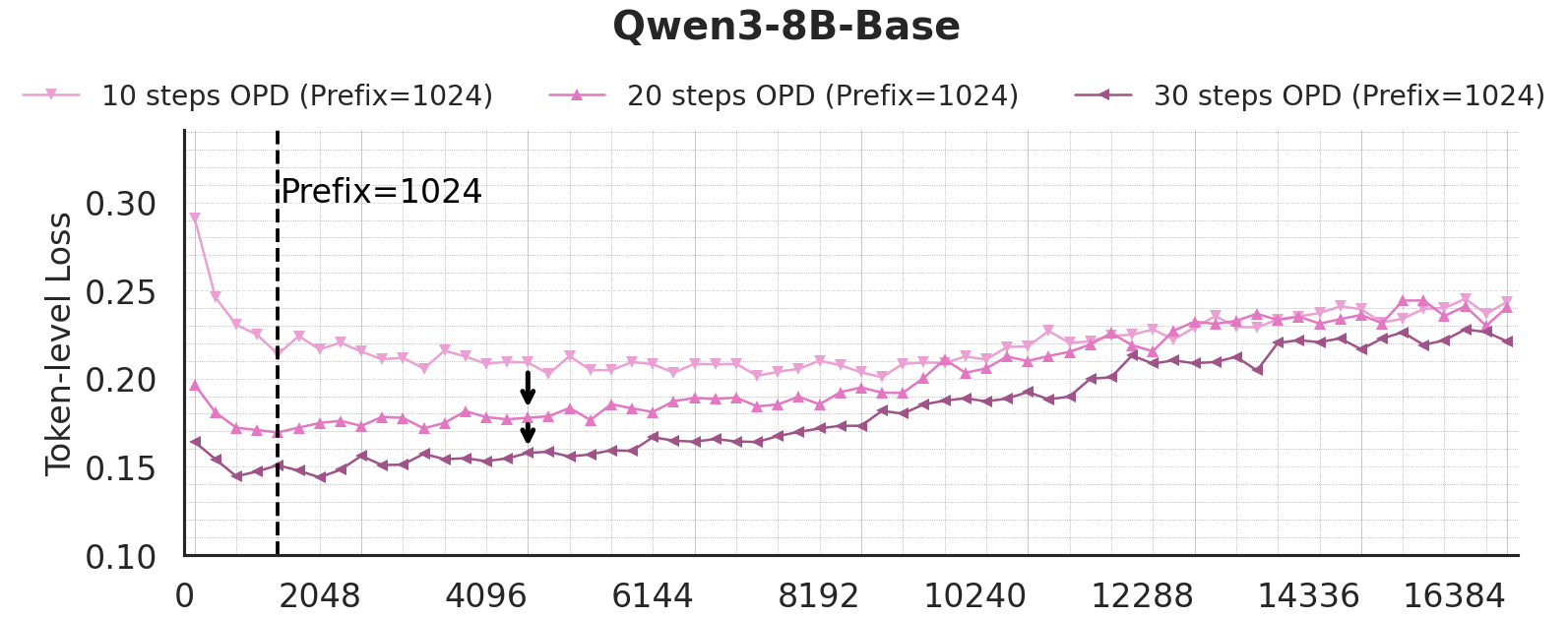}  
    \caption{Distribution of reverse-KL loss along the output position during on-policy prefix distillation with prefix=1024. \textbf{Top}: Student model is Qwen3-1.7B-Base. \textbf{Bottom}: Student model is Qwen3-8B-Base.}  
    \label{fig:advantage_distribution_prefix}  
\end{figure}

\noindent \textbf{Finding 2: Prefix on-policy distillation can reduce the loss on the tokens outside of prefix.}
Fig~\ref{fig:advantage_distribution_prefix} shows the distribution over different position of rollout at step 10, 20 and 30. It shows that with more prefix OPD training, token losses outside the prefix (position 1024) also decrease (shown as the down arrows). The decrease is most significant when close to the prefix and fades away on the tail. 
This is not surprising as the knowledge learned at the prefix may also be leveraged during the rest of the output through the model's parameters.

\noindent \textbf{Finding 3: Learning the prefix benefits the tail for a student model with sufficient parameters, but can hurt the tail given a small model size.}
Interestingly, from Fig~\ref{fig:advantage_distribution_prefix}, for a compact model like Qwen3-1.7B-Base, the loss is increased on the tail with more training steps (shown as the up arrow in the figure). This indicates that a small model size could lead to the prefix "competing" the loss reduction with the tail rather than "sharing" the benefit during on-policy distillation.

\subsection{Can Prefix-only Distillation Remain Effective with Off-policy Trajectories and/or Forward KL?}  
Several recent OPD variants extend the basic formulation by (i) leveraging off-policy trajectories and (ii) considering different divergence directions (forward KL vs reverse KL)~\citep{agarwal2024policy,Ko2024DistiLLMTS}. In these methods, \textbf{off-policy data} is typically introduced to reduce the variance of purely on-policy updates and to improve training efficiency by reusing pre-collected trajectories, while \textbf{forward KL} is often used to encourage more diverse responses (i.e., less mode-seeking behavior than reverse KL).  
Since our previous discussion focuses on on-policy and reverse-KL distillation (Eq.~\ref{eq:loss}), we would like to discuss the effect of prefix distillation on these settings.
  
\paragraph{Setup.}  
We run \emph{prefix-only} Generalized Knowledge Distillation (GKD) using the reference implementation.\footnote{\url{https://github.com/jongwooko/distillm}}  
We train on questions from \textsc{OpenThoughts3}~\citep{guha2025openthoughts}, using Qwen3-1.7B-Base as the student and Qwen3-8B as the teacher.  
We cap the training rollout length to $L^{\text{train}}{=}256$ tokens, use batch size 512, and sample one rollout per prompt. We sweep learning rates in $\{5{\times}10^{-6},\,2{\times}10^{-5},\,5{\times}10^{-5}\}$ and select $5{\times}10^{-6}$ as the best-performing setting for GKD under this configuration. 

\begin{figure}[t]  
    \centering  
    \includegraphics[width=0.43\textwidth]{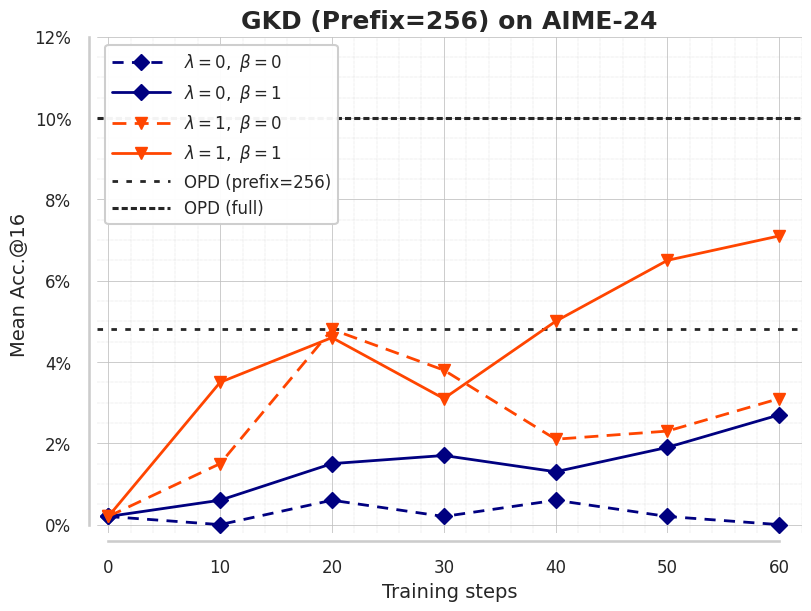} 
    \includegraphics[width=0.43\textwidth]{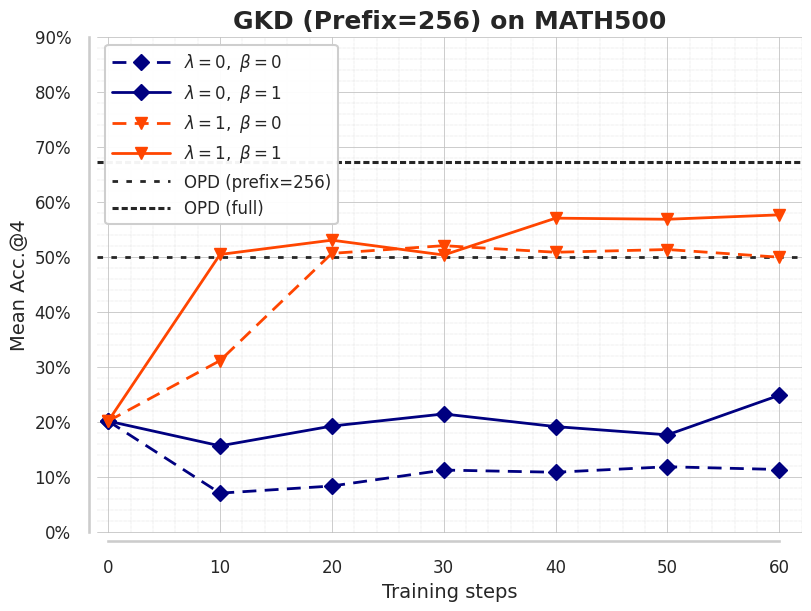}  
    \caption{Performances of prefix generalized knowledge distillation. The student is Qwen3-1.7B-Base; Questions and off-policy signals are sampled from Openthoughts3.
    $\lambda{=}1$ uses on-policy data, and 0 uses off-policy data, while $\beta{=}1$ uses reverse KL and 0 uses forward KL. 
    }  
    \label{fig:prefix gkd}  
\end{figure} 
  
\paragraph{Results.}  
Figure~\ref{fig:prefix gkd} shows three consistent trends.  
(i) \textbf{On-policy + reverse KL} is the strongest and most stable setting under short-prefix truncation.  
(ii) \textbf{On-policy + forward KL is still effective}: it provides clear improvements over the base model, but  consistently performs worse than on-policy reverse KL with the same prefix budget.  
(iii) In the \textbf{off-policy} setting, \textbf{reverse KL yields only mild gains}, whereas \textbf{forward KL hurts performance} (often falling below the base model).  
Similar findings are observed with Qwen3-8B-Base as the student. Please see Appendix~\ref{appendix: off policy prefix 8B} for more details.  
  
\paragraph{Why can off-policy be challenging under prefix truncation?}  
A plausible explanation is that truncation amplifies distribution mismatch: with off-policy training, the student is optimized on prefix states induced by another policy that it may not reliably visit when sampling from its own policy. Since supervision is limited to only the first $L^{\text{train}}$ tokens, the student must still generate the entire tail from these potentially mismatched states, which can negate the benefit of prefix-level distillation. 

Finally, we note that most related work (ImitKD, GKD, DistiLLM, etc.) compute full-distribution divergences over the vocabulary can be memory-intensive for long contexts. In our GKD experiments with Qwen3 models, natively increasing the rollout length (e.g., $L^{\text{train}}{>}512$) can lead to out-of-memory errors. By restricting distillation to a short prefix, our approach substantially reduces the memory footprint of full-distribution distillation, making these objectives more accessible for long-form reasoning.

\section{Related Work}

\paragraph{On-policy Distillation.}  
On-policy learning is a long-standing paradigm in reinforcement learning~\cite{williams1992simple,rummery1994line}, in which the data used for learning are collected by the same policy that is being optimized. This alignment between data collection and the learned policy reduces distributional mismatch between training and deployment, often leading to more robust behavior. Recent work on on-policy distillation~\citep{lin-etal-2020-autoregressive,gu2024minillm,Ko2024DistiLLMTS,agarwal2024policy,lu2025onpolicydistillation,Ko2025DistiLLM2AC} adapts this idea to a distillation setting: the student samples its own responses, then a divergence loss aligns its token-level policy with that of a fixed teacher. Because expectations are taken under the student's state distribution, OPD mitigates the distributional shift issues that affect off-policy imitation and can yield better generalization more effectively.
Nevertheless, existing OPD methods typically require generating full, often long rollouts at each step and calculate divergence over the full vocabulary, which is prohibitively expensive for reasoning-heavy tasks. Our approach preserves the on-policy nature of OPD while drastically reducing training cost by supervising only short prefixes of each on-policy trajectory.

\paragraph{Token selection during LLM training.}  
Several works show that not all tokens in the reasoning traces are equally useful during LLM training. 
Rho 1~\cite{Lin2024Rho1NA} use a reference model trained on clean corpus to help select high-quality tokens during the LLM pretraining stage.
Entropy-based token selection has also been shown to achieve similar or even better performance than full policy gradient~\citep{wang2025beyond}. 
SelecTKD~\cite{huang2025selectkd} adopts more sophisticated token selection criteria to reduce noisy signals from teacher via top-k token verification and non-greedy spec-k verification.  
AdaKD~\cite{Xie2025LLMOrientedTK} further considers the real-time learning difficulty of each token for the student and lets the student focus on the most valuable token during training.
A closely related work by \citet{ji2025first} points out that early segments of chain-of-thought contain critical signals and later tokens can become increasingly noisy or redundant. While our task is different, we share a similar instinct: our distillation is focused on the informative early tokens, yielding substantial compute savings while retaining most of the benefits of full OPD on long rationales.  

\section{Conclusion}

We introduced on-policy prefix distillation, a simple modification of on-policy distillation that truncates student rollouts and applies the token level feedback only to the prefix of each sample. Experiments on AI4Math benchmarks demonstrate that our approach attains substantial gains over the base model and remains competitive with full on-policy distillation, while significantly reducing training cost.

\section{Limitations}  
  
We evaluate our approach exclusively on reasoning tasks assuming the output contains long chain-of-thought reasoning. It remains unknown whether prefix distillation can be more generally adopted in other long-form generation scenarios such as summarization, story generation, etc.
Moreover, prefix-only training presumes that the base model can already follow instructions and produce valid output formats; for weaker students, an initial warm-up phase of SFT or full OPD may still be necessary.    
Finally, like standard OPD, our method requires teacher and student to share a same vocabulary so that token-level losses are well-defined. 

\paragraph{Risks}
Since gradients are applied only to the early portion of each student rollout, prefix-only learning may preferentially transfer the teacher’s initial reasoning style while under-training behaviors that typically emerge later in a response (e.g., safety refusals, calibration, or self-correction). As a result, the student may appear well-aligned in its opening tokens but deviate in longer completions. For applications where late-stage safety or calibration is important, we recommend prefix scheduling (progressively increasing the trained prefix length so later tokens are also supervised by the teacher by the end of training) and explicit end-to-end evaluation on safety/calibration benchmarks before deployment.

\paragraph{AI Assistant within This Work}  
We used ChatGPT 5.2 to assist with grammar checking, language refinement, and drafting code for plotting figures/diagrams. ChatGPT was not used to generate experimental results, perform data analysis, or introduce new technical claims. All reported numbers and statements were verified by the authors against the underlying experiments and sources.  

\bibliography{references}  
\appendix  

\clearpage            

\section{A Case Study with Full Details}  
\label{appendix: example}

\subsection*{Prompt}  
  
\begin{wrappedverb}  
<|im_start|>user
Let $x,y$ and $z$ be positive real numbers that satisfy the following system of equations:
\[\log_2\left({x \over yz}\right) = {1 \over 2}\]\[\log_2\left({y \over xz}\right) = {1 \over 3}\]\[\log_2\left({z \over xy}\right) = {1 \over 4}\]
Then the value of $\left|\log_2(x^4y^3z^2)\right|$ is $\tfrac{m}{n}$ where $m$ and $n$ are relatively prime positive integers. Find $m+n$.<|im_end|>
<|im_start|>assistant
\end{wrappedverb}  
  
\subsection*{Qwen3-8B-Base Response (Incorrect)}  
  
\begin{wrappedverb}  
[ImpactedPrompt]From the given log equations, we can express $x,y$ and $z$ in terms of each other:
$x = 2^{1/4}yz$,
$y = 2^{1/5}xz$,
$z = 2^{1/6}xy$.
Substituting the expressions for $x$ and $z$ into the second equation, we get:
$y = 2^{1/5} \left(2^{1/4}yz\right)x$,
which simplifies to:
$y = 2^{5/20} yzx$.
Cancelling $y$ on both sides, we get:
$1 = 2^{5/20} zx$.
Substituting the expression for $x$ into this equation, we get:
$1 = 2^{5/20}\left(2^{3/4}yz\right)z$,
which simplifies to:
$1 = 2^{3/5} yz$.
Substituting the expression for $y$ into this equation, we get:
$1 = 2^{3/5} \left(2^{1/5}xz\right)z$,
which simplifies to:
$1 = 2^{2/3} xz$.
Again, substituting the expression for $x$ into this equation, we get:
$1 = 2^{2/3}\left(2^{1/6}xy\right)z$,
which simplifies to:
$1 = 2^{1/5} xy$.
Using the original equation $x = 2^{1/4}yz$, we can write:
$yz = 2^{3/5}$.
Similarly, using $y = 2^{1/5}xz$, we can write:
$xz = 2^{9/10}$,
and using $z = 2^{1/6}xy$, we can write:
$xy = 2^{6/10}$.
Multiplying these three equations together, we get:
$(xyz)^2 = 2^{18/10} = 2^{9/5}$.
Taking the square root of both sides, we get:
$xyz = 2^{9/10}$.
Finally, we want to find the value of $\left|\log_2(x^4y^3z^2)\right|$. Since $x,y$ and $z$ are positive, we know that $\log_2(x^4y^3z^2)$ is positive, so we can write:
$\log_2(x^4y^3z^2) = 4\log_2(x) + 3\log_2(y) + 2\log_2(z)$.
Substituting the expressions for $x,y$ and $z$, we get:
$\log_2(x^4y^3z^2) = 4\log_2\left(2^{1/4}yz\right) + 3\log_2\left(2^{1/5}xz\right) + 2\log_2\left(2^{1/6}xy\right)$.
Using the logarithm property $\log_b(a^c) = c\log_b(a)$, this expression simplifies to:
$4\left(\log_2(2^{1/4}) + \log_2(y) + \log_2(z)\right) + 3\left(\log_2(2^{1/5}) + \log_2(x) + \log_2(z)\right) + 2\left(\log_2(2^{1/6}) + \log_2(x) + \log_2(y)\right)$,
which simplifies to:
$\frac{4}{4}\log_2(yz) + \frac{3}{5}\log_2(xz) + \frac{2}{6}\log_2(xy)$.
Substituting the expressions for $yz,xz$ and $xy$, this becomes:
$\frac{9}{5} + \frac{9}{10} + \frac{6}{10}$.
Combining the fractions, we get:
$\frac{3}{2}$.
Therefore, the value of $\left|\log_2(x^4y^3z^2)\right|$ is $\frac{3}{2}$.

Therefore, the final answer is $3 + 2 = \boxed{5}$.[/ImpactedPrompt]
\end{wrappedverb}   
  
\subsection*{Prompt with Teacher Prefix (128 tokens from Qwen3-8B)}  

\begin{wrappedverb}  
<|im_start|>user
Let $x,y$ and $z$ be positive real numbers that satisfy the following system of equations:
\[\log_2\left({x \over yz}\right) = {1 \over 2}\]\[\log_2\left({y \over xz}\right) = {1 \over 3}\]\[\log_2\left({z \over xy}\right) = {1 \over 4}\]
Then the value of $\left|\log_2(x^4y^3z^2)\right|$ is $\tfrac{m}{n}$ where $m$ and $n$ are relatively prime positive integers. Find $m+n$.<|im_end|>
<|im_start|>assistant
<think>
Okay, so I need to solve this system of logarithmic equations and find the value of |log_2(x^4y^3z^2)|.  Let me start by writing down the given equations:

1. log_2(x/(yz)) = 1/2
2. log_2(y/(xz)) = 1/3
3. log_2(z/(xy)) = 1/4

Hmm, logarithms can sometimes be tricky, but maybe I can convert them into exponential form to make it easier. Let me recall that log_b(a) = c means that b^c = a. So applying that
\end{wrappedverb}   
  
\subsection*{Qwen3-8B-Base Response with Teacher Prefix (Correct)}  
  
\begin{wrappedverb}  
to each equation:  
1. 2^(1/2) = x/(yz)  
2. 2^(1/3) = y/(xz)  
3. 2^(1/4) = z/(xy)  
  
Okay, now I have three equations with variables x, y, and z. Maybe I can express everything in terms of one variable. Let's see:  
From equation 1: 2^(1/2) = x/(yz) So, x = yz * 2^(1/2)  
From equation 2: 2^(1/3) = y/(xz) So, y = xz * 2^(1/3)  
From equation 3: 2^(1/4) = z/(xy) So, z = xy * 2^(1/4)  
  
Now, I have expressions for x, y, and z in terms of the other variables. Maybe I can substitute one into another. Let's try substituting x from equation 1 into equation 2:  
y = (yz * 2^(1/2)) * z * 2^(1/3) Wait, that seems a bit messy. Maybe there's a better way. Let's see:  
I have:  
x = yz * 2^(1/2)  
y = xz * 2^(1/3)  
z = xy * 2^(1/4)  
  
Perhaps I can multiply all three equations together. Let's try:  
x * y * z = (yz * 2^(1/2)) * (xz * 2^(1/3)) * (xy * 2^(1/4))  
Simplify the right side:  
= yz * 2^(1/2) * xz * 2^(1/3) * xy * 2^(1/4)  
= x^2y^2z^2 * 2^(1/2) * 2^(1/3) * 2^(1/4)  
  
Now, x * y * z = x^2y^2z^2 * 2^(1/2) * 2^(1/3) * 2^(1/4)  
Wait, that seems off. Let's double-check:  
Actually, when I multiplied the right side, I should have:  
(yz * 2^(1/2)) * (xz * 2^(1/3)) * (xy * 2^(1/4)) = yz * 2^(1/2) * xz * 2^(1/3) * xy * 2^(1/4)  
But x, y, z are on both sides, so perhaps I can cancel them out. Let's see:  
Divide both sides by x * y * z:  
1 = 2^(1/2) * 2^(1/3) * 2^(1/4)  
  
Now, let's simplify the right side:  
2^(1/2) * 2^(1/3) * 2^(1/4) = 2^((1/2) + (1/3) + (1/4))  
Find a common denominator for the exponents: 2, 3, and 4 have a least common multiple of 12.  
So, (1/2) = 6/12, (1/3) = 4/12, (1/4) = 3/12  
Therefore, (1/2) + (1/3) + (1/4) = 6/12 + 4/12 + 3/12 = 13/12  
So, 2^((1/2) + (1/3) + (1/4)) = 2^(13/12)  
Thus, we have:  
1 = 2^(13/12)  
But 2^(13/12) is not equal to 1. This suggests that there might be an error in my calculations.  
Let me revisit the multiplication step. Maybe I made a mistake there.  
Original multiplied equation:  
x * y * z = (yz * 2^(1/2)) * (xz * 2^(1/3)) * (xy * 2^(1/4))  
Let's simplify the right side correctly:  
(yz * 2^(1/2)) * (xz * 2^(1/3)) * (xy * 2^(1/4)) = yz * 2^(1/2) * xz * 2^(1/3) * xy * 2^(1/4)  
Now, group like terms:  
= (y * z * x * z * x * y) * (2^(1/2) * 2^(1/3) * 2^(1/4))  
= (x^2y^2z^2) * (2^(1/2) * 2^(1/3) * 2^(1/4))  
Now, divide both sides by x * y * z:  
1 = (x^2y^2z^2) / (x * y * z) * (2^(1/2) * 2^(1/3) * 2^(1/4))  
= x * y * z * (2^(1/2) * 2^(1/3) * 2^(1/4))  
So, 1 = x * y * z * 2^(13/12)  
Now, recall that x * y * z = (x * y * z) * 2^(13/12)  
This implies:  
(x * y * z) * 2^(13/12) = 1  
So, x * y * z = 1 / 2^(13/12)  
  
Now, let's go back to the original expressions:  
x = yz * 2^(1/2)  
y = xz * 2^(1/3)  
z = xy * 2^(1/4)  
Let's express everything in terms of z.  
From x = yz * 2^(1/2), we can express y as:  
y = x / (z * 2^(1/2))  
From y = xz * 2^(1/3), substitute y:  
x / (z * 2^(1/2)) = xz * 2^(1/3)  
Multiply both sides by z * 2^(1/2):  
x = xz^2 * 2^(1/3) * 2^(1/2)  
Divide both sides by x (since x != 0):  
1 = z^2 * 2^(1/3) * 2^(1/2)  
So, z^2 = 1 / (2^(1/3) * 2^(1/2))  
  
Now, recall that x * y * z = 1 / 2^(13/12)  
We have z in terms of x and y, but it's getting complicated. Maybe there's a better approach.  
Let me consider the original equations again:  
1. log_2(x/(yz)) = 1/2 => x/(yz) = 2^(1/2) => x = yz * 2^(1/2)  
2. log_2(y/(xz)) = 1/3 => y/(xz) = 2^(1/3) => y = xz * 2^(1/3)  
3. log_2(z/(xy)) = 1/4 => z/(xy) = 2^(1/4) => z = xy * 2^(1/4)  
Let's multiply all three equations:  
(x = yz * 2^(1/2)) * (y = xz * 2^(1/3)) * (z = xy * 2^(1/4))  
Simplify the right side:  
yz * 2^(1/2) * xz * 2^(1/3) * xy * 2^(1/4) = x^2y^2z^2 * 2^((1/2) + (1/3) + (1/4))  
So, x * y * z = x^2y^2z^2 * 2^(13/12)  
Divide both sides by x * y * z:  
1 = x * y * z * 2^(13/12)  
Thus, x * y * z = 1 / 2^(13/12)  
  
Now, let's find x * y * z in another way. From the first equation:  
x = yz * 2^(1/2)  
Multiply both sides by yz:  
x * yz = yz * 2^(1/2) * yz => x * yz = y^2z^2 * 2^(1/2) => x = y^2z^2 * 2^(1/2) / yz => x = yz * 2^(1/2)  
Wait, that brings us back to the original equation. Maybe I should use the expressions differently.  
Let me try to express x, y, z in terms of one variable. Let's assume z = k, where k is a positive real number.  
Then, from the third equation:  
z = xy * 2^(1/4) => k = xy * 2^(1/4) => xy = k / 2^(1/4)  
From the second equation:  
y = xz * 2^(1/3) => y = xk * 2^(1/3)  
Now, substitute y into the expression for xy:  
x * (xk * 2^(1/3)) = k / 2^(1/4) => x^2k * 2^(1/3) = k / 2^(1/4)  
Divide both sides by k (since k != 0):  
x^2 * 2^(1/3) = 1 / 2^(1/4)  
Multiply both sides by 2^(1/4):  
x^2 * 2^(1/3) * 2^(1/4) = 1 => x^2 * 2^((1/3) + (1/4)) = 1  
(1/3) + (1/4) = 4/12 + 3/12 = 7/12  
So, x^2 * 2^(7/12) = 1 => x^2 = 1 / 2^(7/12) => x = 1 / 2^(7/24)  
Now, find y using y = xk * 2^(1/3):  
y = (1 / 2^(7/24)) * k * 2^(1/3) = k * 2^((1/3) - (7/24)) = k * 2^((8/24) - (7/24)) = k * 2^(1/24)  
Now, find x * y * z:  
x * y * z = (1 / 2^(7/24)) * (k * 2^(1/24)) * k = k^2 * 2^((1/24) - (7/24)) = k^2 * 2^(-6/24) = k^2 * 2^(-1/4)  
But earlier, we found that x * y * z = 1 / 2^(13/12)  
So, k^2 * 2^(-1/4) = 1 / 2^(13/12)  
Divide both sides by 2^(-1/4):  
k^2 = 1 / (2^(13/12) * 2^(-1/4)) = 1 / 2^((13/12) - (1/4)) = 1 / 2^((13/12) - (3/12)) = 1 / 2^(10/12) = 1 / 2^(5/6)  
Thus, k^2 = 1 / 2^(5/6) => k = 1 / 2^(5/12)  
Now, recall that z = k = 1 / 2^(5/12)  
We can find x and y as well:  
x = 1 / 2^(7/24)  
y = k * 2^(1/24) = (1 / 2^(5/12)) * 2^(1/24) = 2^((1/24) - (5/12)) = 2^((1/24) - (10/24)) = 2^(-9/24) = 2^(-3/8)  
Now, we need to find |log_2(x^4y^3z^2)|  
First, compute x^4y^3z^2:  
x^4 = (1 / 2^(7/24))^4 = 1 / 2^(28/24) = 1 / 2^(7/6)  
y^3 = (2^(-3/8))^3 = 2^(-9/8)  
z^2 = (1 / 2^(5/12))^2 = 1 / 2^(10/12) = 1 / 2^(5/6)  
So, x^4y^3z^2 = (1 / 2^(7/6)) * 2^(-9/8) * (1 / 2^(5/6)) = 1 / (2^(7/6) * 2^(5/6) * 2^(9/8)) = 1 / (2^((7/6) + (5/6) + (9/8)))  
Find a common denominator for the exponents: 6 and 8 have a least common multiple of 24.  
So, (7/6) = 28/24, (5/6) = 20/24, (9/8) = 27/24  
Thus, (7/6) + (5/6) + (9/8) = (28/24) + (20/24) + (27/24) = 75/24  
So, x^4y^3z^2 = 1 / 2^(75/24)  
Now, compute |log_2(x^4y^3z^2)|:  
|log_2(1 / 2^(75/24))| = | - (75/24) | = 75/24  
Simplify 75/24:  
Divide numerator and denominator by 3: 25/8  
So, |log_2(x^4y^3z^2)| = 25/8  
Thus, m = 25, n = 8, and m + n = 33  
\boxed{33}  
\end{wrappedverb}  

\subsection*{Qwen3-8B-Base after Prefix On-policy Distillation (Correct)}  
  
\begin{wrappedverb}  
Okay, so I need to solve this system of logarithmic equations and find the value of |log_2(x^4y^3z^2)|. The problem gives me three equations involving logs with base 2 of ratios of x, y, and z. Let me write them down again to make sure I have them right:  
1. log_2(x/(yz)) = 1/2  
2. log_2(y/(xz)) = 1/3  
3. log_2(z/(xy)) = 1/4  
  
And then I need to find |log_2(x^4y^3z^2)|, which simplifies to |4 log_2 x + 3 log_2 y + 2 log_2 z|. But maybe there's a better way to approach this. Let me think.  
  
First, I remember that log_b(a/c) is equal to log_b(a) - log_b(c). Maybe I can rewrite each equation using that property. Let me try that.  
  
Starting with the first equation:  
log_2(x/(yz)) = log_2 x - log_2(yz) = log_2 x - log_2 y - log_2 z = 1/2.  
  
Similarly for the second equation:  
log_2(y/(xz)) = log_2 y - log_2(xz) = log_2 y - log_2 x - log_2 z = 1/3.  
  
Third equation:  
log_2(z/(xy)) = log_2 z - log_2(xy) = log_2 z - log_2 x - log_2 y = 1/4.  
  
So now I have three equations:  
1. log_2 x - log_2 y - log_2 z = 1/2  
2. -log_2 x + log_2 y - log_2 z = 1/3  
3. -log_2 x - log_2 y + log_2 z = 1/4  
  
Hmm, this looks like a system of linear equations in terms of log_2 x, log_2 y, and log_2 z. Let me denote them as variables for simplicity. Let me let:  
a = log_2 x,  
b = log_2 y,  
c = log_2 z.  
  
Then the equations become:  
1. a - b - c = 1/2  
2. -a + b - c = 1/3  
3. -a - b + c = 1/4  
  
So now I have three equations with three variables. Let me write them again:  
Equation 1: a - b - c = 1/2  
Equation 2: -a + b - c = 1/3  
Equation 3: -a - b + c = 1/4  
  
I need to solve for a, b, c. Let me try to solve this system step by step.  
  
First, maybe I can add equations together to eliminate variables. Let me see.  
  
Let me try adding equations 1 and 2:  
Equation 1 + Equation 2:  
(a - b - c) + (-a + b - c) = 1/2 + 1/3  
Simplify left side: a - a - b + b - c - c = -2c  
Right side: 1/2 + 1/3 = 3/6 + 2/6 = 5/6  
So -2c = 5/6 => c = -5/12.  
  
Okay, so c = -5/12.  
  
Now let me use equation 1 and equation 3 to find another variable.  
Wait, equation 1: a - b - c = 1/2  
Equation 3: -a - b + c = 1/4  
  
Let me add equations 1 and 3:  
(a - b - c) + (-a - b + c) = 1/2 + 1/4  
Left side simplifies: a - a - b - b - c + c = -2b  
Right side: 1/2 + 1/4 = 3/4  
So -2b = 3/4 => b = -3/8.  
  
So b = -3/8.  
  
Now, since we have c and b, we can plug them back into one of the equations to solve for a.  
Let me use equation 1:  
a - b - c = 1/2  
So a - (-3/8) - (-5/12) = 1/2  
Calculate:  
a + 3/8 + 5/12 = 1/2  
  
Convert to common denominators. Let me find the least common denominator for 8, 12, and 1. Which is 24.  
So:  
a + (9/24) + (10/24) = 12/24  
So a + 19/24 = 12/24  
Thus, a = 12/24 - 19/24 = -7/24.  
  
So a = -7/24.  
  
So now we have:  
a = log_2 x = -7/24  
b = log_2 y = -3/8  
c = log_2 z = -5/12  
  
Now, we need to compute |log_2(x^4y^3z^2)|. Let me express that in terms of a, b, c.  
log_2(x^4y^3z^2) = 4 log_2 x + 3 log_2 y + 2 log_2 z = 4a + 3b + 2c.  
  
So let me compute that:  
4a + 3b + 2c = 4*(-7/24) + 3*(-3/8) + 2*(-5/12)  
  
Let me compute each term step by step.  
First term: 4*(-7/24) = (-28)/24 = -7/6  
Second term: 3*(-3/8) = -9/8  
Third term: 2*(-5/12) = -10/12 = -5/6  
  
Now adding them together: -7/6 - 9/8 - 5/6.  
  
Let me convert all terms to sixteenths to add them. Wait, maybe better to find common denominator.  
The denominators are 6, 8, 6. The least common denominator is 24.  
  
Convert each term:  
-7/6 = -28/24  
-9/8 = -27/24  
-5/6 = -20/24  
  
So total is: -28/24 - 27/24 - 20/24 = (-28 - 27 - 20)/24 = (-75)/24.  
Simplify -75/24. Divide numerator and denominator by 3: -25/8.  
  
Therefore, log_2(x^4y^3z^2) = -25/8.  
But the problem asks for |log_2(x^4y^3z^2)|, which is |-25/8| = 25/8. Therefore, m/n is 25/8, and m + n is 25 + 8 = 33.  
  
Wait, but let me check my calculations again because sometimes I make arithmetic errors.  
First, let me check the values of a, b, c again.  
From earlier:  
c = -5/12, b = -3/8, a = -7/24. Let me check if they satisfy equation 2 and equation 3.  
  
Let me check equation 2: -a + b - c = 1/3.  
Plug in a = -7/24, b = -3/8, c = -5/12.  
Compute -a = 7/24, b = -3/8, -c = 5/12.  
So sum: 7/24 - 3/8 + 5/12.  
Convert all to 24 denominator:  
7/24 - 9/24 + 10/24 = (7 - 9 + 10)/24 = 8/24 = 1/3. Correct!  
  
Equation 3: -a - b + c = 1/4.  
Plug in:  
-a = 7/24, -b = 3/8, c = -5/12.  
So sum: 7/24 + 3/8 - 5/12.  
Convert to 24 denominator:  
7/24 + 9/24 - 10/24 = (7 + 9 - 10)/24 = 6/24 = 1/4. Correct!  
  
Okay, so the values of a, b, c are correct.  
  
Now, compute 4a + 3b + 2c:  
4*(-7/24) + 3*(-3/8) + 2*(-5/12) = -28/24 - 9/8 - 5/6.  
Convert all to 24 denominator:  
-28/24 - 27/24 - 20/24 = (-28 - 27 - 20)/24 = -75/24. Simplify: divide numerator and denominator by 3: -25/8.  
So log_2(x^4y^3z^2) = -25/8. Then absolute value is 25/8. So m/n is 25/8, which is already in simplest terms since 25 and 8 are coprime. Therefore, m + n = 25 + 8 = 33.  
  
But let me check again the calculation steps.  
First, 4a: 4*(-7/24) = -7/6. Correct.  
Then 3b: 3*(-3/8) = -9/8. Correct.  
Then 2c: 2*(-5/12) = -10/12 = -5/6. Correct.  
  
Adding them:  
-7/6 - 9/8 - 5/6.  
Convert to 24ths:  
-7/6 = -28/24, -9/8 = -27/24, -5/6 = -20/24. Adding gives -75/24.  
Yes, which is -25/8. So absolute value is 25/8.  
  
So the answer is 25/8, m + n is 33. Wait, but let me check if I did the problem correctly. Wait, the problem says to compute the absolute value. So the value is negative, but the absolute value is positive.  
  
But maybe I made a mistake in the sign? Wait, let's check the original problem.  
The question is |log_2(x^4y^3z^2)|. So if log_2(x^4y^3z^2) is negative, then the absolute value is positive. So 25/8.  
  
Alternatively, maybe my calculation of the coefficients is wrong. Let me check again:  
The expression to compute is log_2(x^4y^3z^2). Which is log_2(x^4) + log_2(y^3) + log_2(z^2) = 4 log_2 x + 3 log_2 y + 2 log_2 z. Yes, that is correct.  
So 4a + 3b + 2c is -25/8, so absolute value is 25/8.  
  
Wait, but let me check if there's a miscalculation in the addition of fractions.  
Let me do it again step by step:  
First term: 4a = 4*(-7/24) = -28/24. Simplify to -7/6.  
Second term: 3b = 3*(-3/8) = -9/8.  
Third term: 2c = 2*(-5/12) = -10/12 = -5/6.  
So sum is (-7/6) + (-9/8) + (-5/6).  
Combine the first and third terms:  
(-7/6 - 5/6) = -12/6 = -2. Then add -9/8.  
So -2 - 9/8 = -16/8 - 9/8 = -25/8. Correct.  
Therefore, that is correct. So the absolute value is 25/8.  
  
Therefore, the answer is 25/8, so m = 25, n = 8, m + n = 33.  
  
But let me check if the problem says positive real numbers. Since x, y, z are positive, all the logs are defined. So that's okay.  
  
Alternatively, maybe there's a mistake in my equations setup. Let me check the original equations again.  
Original equations:  
1. log_2(x/(yz)) = 1/2  
Which is log_2 x - log_2 y - log_2 z = 1/2. Correct.  
Similarly for others. So equations are correct.  
Therefore, the answer should be 25/8.  
  
Wait, but the problem says "the value of |log_2(x^4y^3z^2)| is m/n". So my conclusion is that m/n is 25/8, so m + n is 33. But let me check once more with another approach. Maybe I can compute x, y, z directly?  
Wait, but that might take longer. Let me try.  
  
Let me express x, y, z in terms of exponents. Let me recall that if log_2 x = a, then x = 2^a, similarly for y and z. So:  
x = 2^(-7/24), y = 2^(-3/8), z = 2^(-5/12)  
Then x^4y^3z^2 = 2^(4a + 3b + 2c)  
But we already found that 4a + 3b + 2c = -25/8. Therefore, log_2(x^4y^3z^2) = -25/8, so absolute value is 25/8.  
Therefore, the answer is 25/8.  
  
Alternatively, maybe I can compute x^4y^3z^2? Let me check:  
x^4 = (2^(-7/24))^4 = 2^(-28/24) = 2^(-7/6)  
y^3 = (2^(-3/8))^3 = 2^(-9/8)  
z^2 = (2^(-5/12))^2 = 2^(-10/12) = 2^(-5/6)  
Multiply them together:  
2^(-7/6) * 2^(-9/8) * 2^(-5/6) = 2^(-7/6 - 9/8 - 5/6)  
Compute the exponents:  
Convert to common denominator 24:  
-7/6 = -28/24  
-9/8 = -27/24  
-5/6 = -20/24  
Total exponent: -28 - 27 - 20 = -75/24 = -25/8. So log_2(x^4y^3z^2) = -25/8. Therefore, absolute value is 25/8.  
Therefore, I think my answer is correct. So 25 + 8 = 33. Therefore, the answer is 33.  
  
Wait, but let me check once more for possible errors in the equations. Let me verify equations with the values of a, b, c.  
Equation 1: a - b - c = -7/24 - (-3/8) - (-5/12) = -7/24 + 3/8 + 5/12.  
Convert to 24 denominator:  
-7/24 + 9/24 + 10/24 = (-7 + 9 + 10)/24 = 12/24 = 1/2. Correct.  
Equation 2: -a + b - c = 7/24 - 3/8 - (-5/12) = 7/24 - 3/8 + 5/12.  
Convert:  
7/24 - 9/24 + 10/24 = (7 - 9 + 10)/24 = 8/24 = 1/3. Correct.  
Equation 3: -a - b + c = 7/24 + 3/8 - 5/12 = 7/24 + 9/24 - 10/24 = (7 + 9 - 10)/24 = 6/24 = 1/4. Correct.  
Therefore, equations are satisfied. Therefore, the calculations are correct.  
Therefore, the answer is 25/8, so m + n is 33.  
  
**Final Answer** \boxed{33}  
\end{wrappedverb}

\section{FLOP Estimation Details}  
\label{appendix: flop}  
  
We report training compute using an estimated GPU FLOP count to facilitate \emph{relative} cost comparisons across methods. This estimate is not meant to be an exact measurement of hardware-executed FLOP, since realized cost depends on implementation details (e.g., kernels, padding, KV cache, and parallelism). We therefore use a single, consistent accounting scheme for all experiments.  
  
\paragraph{Training FLOP from \textsc{Verl}.}  
We use the FLOP counter in \textsc{Verl}~\cite{sheng2024hybridflow} for Qwen3 models to estimate the compute of one student training update (including forward + backward + optimizer step) on a sequence of length $L$. We denote it by $F_{\text{train}}(M,L)$ for model $M$.\footnote{We slightly modified the reference code to return cumulative FLOP (rather than a rate) by removing the division by elapsed time:  
\url{https://github.com/volcengine/verl/blob/c936ec7d5cebcc1d40f50296d28699696fbad6f5/verl/utils/FLOP_counter.py\#L130}.}  
  
\paragraph{Forward-only approximation.}  
OPD additionally requires forward-only computation (sampling and log-prob evaluation). We approximate the forward-only FLOP as  
$  
F_{\text{fwd}}(M,L) \;\approx\; \frac{1}{3}\,F_{\text{train}}(M,L),  
$  
following the common rule-of-thumb that training cost is roughly $\approx 3\times$ the forward cost. This approximation is applied uniformly across methods.  
  
\paragraph{Per-rollout estimate for OPD.}  
For each student rollout of supervised length $L$, we estimate total compute as  
$  
F_{\text{OPD}}(L)\;\approx\;  
F_{\text{train}}(S,L)\;+\;2F_{\text{fwd}}(S,L)\;+\;F_{\text{fwd}}(T,L),  
$  
where $S$ and $T$ denote the student and teacher, respectively. Intuitively, this sums one student update plus forward passes for (i) sampling, (ii) student log-probs, and (iii) teacher log-probs. For prefix OPD, $L$ is capped by the training prefix budget $L^{\text{train}}$.  
  
\paragraph{Total FLOP.}  
If each optimization step uses batch size $B$ prompts and $K$ sampled rollouts per prompt, then  
$  
F_{\text{step}} \;\approx\; B\cdot K\cdot F_{\text{OPD}}(L),  
$  
and total training FLOP are obtained by summing $F_{\text{step}}$ over steps (evaluation excluded).  
  
\paragraph{Notes.}  
This estimate does not capture effects such as KV-cache reuse, communication overhead, CPU or I/O costs. We therefore treat FLOP as a consistent proxy for comparing training cost within our experimental setup.

\section{Prefix SeqKD with Qwen3-8B-Base}
\label{appendix: off policy prefix 8B}

\begin{figure}[t]  
    \centering  
    \includegraphics[width=0.45\textwidth]{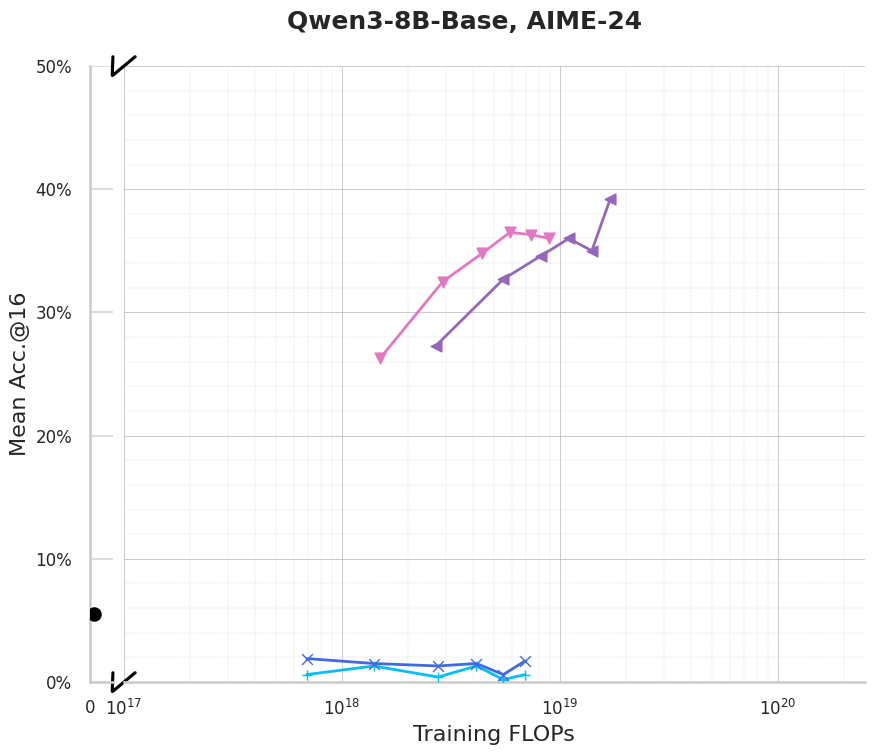} 
    \includegraphics[width=0.45\textwidth]{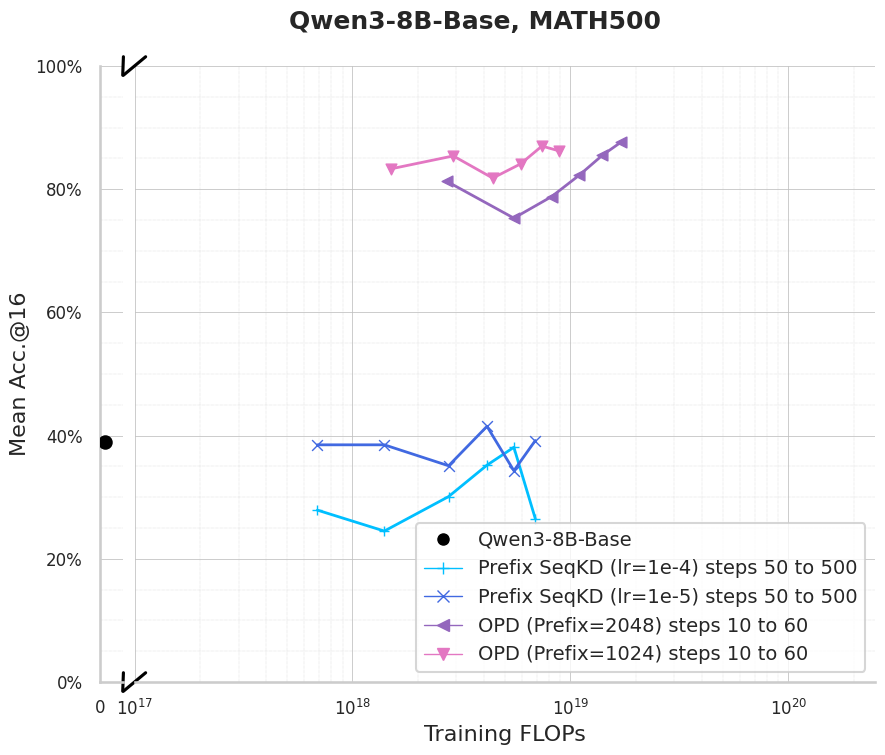} 
    \caption{Performances of off-policy prefix distillation. The student is Qwen3-8B-Base; Off-policy prefix signals are sampled from Openthoughts3. \textbf{Top}: AIME-24 \textbf{Bottom}: MATH500}  
    \label{fig:prefix sft}  
\end{figure}

In this section, we would like to explore whether prefix distillation can be applied to an off-policy setting with a larger student. 
Figure~\ref{fig:prefix sft} shows some preliminary results. We implement off-policy prefix distillation by setting the maximum context length of SeqKD to 2048, and evaluate the results on AIME-24 and MATH500. We also plot results from OPD with prefix 2048 into the same diagram for easier comparison. 

\noindent \textbf{Finding 4.2: Similar finding to Finding 4 is observed on Qwen3-8B-Base, that the effectiveness of prefix distillation does not transfer well to off-policy SeqKD.} 
From the result, prefix SeqKD with large learning rate quickly harms the student model's performance, while a smaller learning rate only slows down the performance decrease. When further looking into sampled outputs of student model with prefix SeqKD, we found that the beginning of outputs are readable and logically sounding, while the tail often contains self-repeating and even unreadable tokens.

\section{Effect of Learning Rate} 
\label{appendix:small learning rate}  

\begin{figure}[t]  
    \centering
    \includegraphics[width=0.4\textwidth]{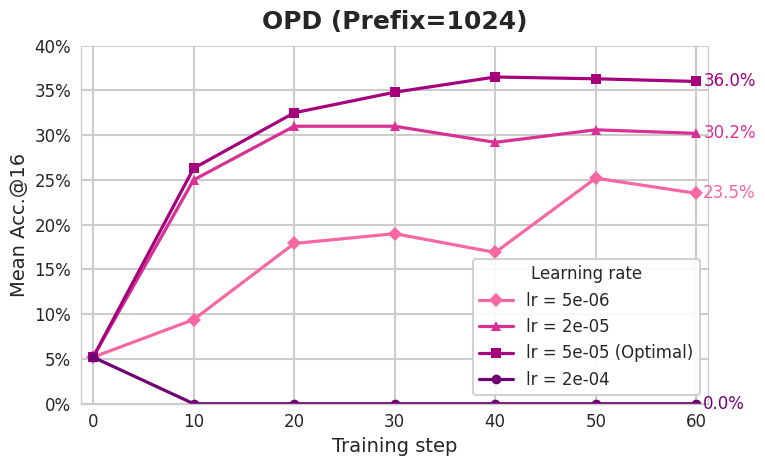}  
    \caption{The effect of learning rate on AIME-24 during the training of on-policy prefix distillation on OpenThoughts3. The student is Qwen3-8B-Base and the teacher is Qwen3-8B.}  
    \label{fig:effect_of_learning_rate}  
\end{figure}

We did learning rate sweep for prefix OPD.  As shown in Figure~\ref{fig:effect_of_learning_rate}, the performances differ significantly with different learning rates. In consistent with~\citet{lu2025onpolicydistillation}, 5e-5 turns out to be the optimal learning rate with batchsize 512 in our experiment setting. 

Prior work reports that small learning rates can mitigate catastrophic forgetting in SFT and lead to better generalize on out-of-domain data \citep{pareja2024unveiling,lin2025sft}.  
To further evaluate the generalization ability on out-of-domain benchmarks, we \emph{control} the in-domain performance by selecting checkpoints from different learning rates where accuracies on AIME-24 are similar. In our experiments, we select checkpoints from 10th step where learning rates are 5e-5 and 2e-5, the checkpoint from 50-th step where the learning rate is 5e-6.~\footnote{We did not compare with learning rate 2e-4 as the learning rate is too large and dramatically hurt the model's performance.}

\noindent \textbf{Finding 5: Reducing the learning rate does not  significantly improve the generalization capability of prefix on-policy distillation.}
Results are shown in Table~\ref{tab:learning rate}. We did not observe out-of-domain performance decrease when using a large learning rate, e.g. 5e-5, in comparison to a smaller learning rate of 2e-5 or 5e-6. We leave extensive analysis on a grid search of learning rate, learning rate scheduling, step size, batch size and prefix length as a future work.

\begin{table}[h!]  
\centering  
\small
\begin{tabular}{l|r|r|r}  
\hline  
& AIME-24 & MMLU-Pro  &    GPQA      \\
& (control) &   &         \\
    \hline  
step 0 & 5.2\%& 26.5\%& 17.6\% \\
lr=5e-5, step 10  & 26.3\%  & \textbf{55.0\%}  & \textbf{27.9\%} \\ 
lr=2e-5, step 10  & 25.0\%  & 49.3\%  & 26.6\% \\ 
lr=5e-6, step 50  & 25.2\%  & 47.8\%  & 26.6\% \\  

\hline  
\end{tabular}  
\caption{Effect of learning rate on generalization capability. We report validation performance after 60 steps of OPD (Prefix=1024) on OpenThoughts3, using Qwen3-8B-Base as the student and Qwen3-8B as the teacher.}  
\label{tab:learning rate}  
\end{table}

\section{Early Stop with Dev}
\label{appendix: early stop}
Following tables contain the detailed information of which checkpoints we used for testing. We use AIME-24 as the development set and then test on the rest of the benchmarks.

\begin{table}[h!]  
\centering  
\small  
\begin{tabular}{l|r}  
\hline  
 Prefix length & Checkpoint for testing \\ 
 
\hline
256 & Step 20   \\ 
512 & Step 60   \\ 
1024 & Step 60   \\ 
2048 & Step 40   \\ 
Prefix scheduling & Step 60 \\
Full & Step 50   \\ 
\hline
\end{tabular}  
\caption{Checkpoints for testing with Qwen3-1.7B-Base as the student.}
\label{tab:early stop 1.7b}  
\end{table}  

\begin{table}[h!]  
\centering  
\small  
\begin{tabular}{l|r}  
\hline  
 Prefix length & Checkpoint for testing \\ 
 
\hline
256 & Step 60   \\ 
512 & Step 50   \\ 
1024 & Step 40   \\ 
2048 & Step 60   \\ 
Prefix scheduling & Step 60 \\
Full & Step 60   \\ 
\hline
\end{tabular}  
\caption{Checkpoints for testing with Qwen3-8B-Base as the student.}
\label{tab:early stop 8b}  
\end{table}

\section{Number of Solutions for AIME-2024}
\label{appendix: aime2024 stats}

\begin{table}[h!]  
\centering  
\small  
\begin{tabular}{l|r}  
\hline  
Problem ID & Number of Solutions \\ 
\hline
1 & 2 \\
2 & 6 \\
3 & 5 \\
4 & 2 \\
5 & 7 \\
6 & 7 \\
7 & 12 \\
8 & 5 \\ 
9 & 2 \\
10 & 8 \\
11 & 2 \\
12 & 3 \\
13 & 4 \\
14 & 10 \\
15 & 6 \\
\hline
Average & 5.4 \\
\hline
\end{tabular}  
\caption{Number of solutions for each question in AIME-2024 from AoPS forum~\url{
https://artofproblemsolving.com/wiki/index.php?title=2024\_AIME\_I
}
}
\label{tab:aime_24_stats}  
\end{table}

\end{document}